\documentclass[10pt,twocolumn,letterpaper]{article}

\usepackage{cvpr} %
\usepackage[accsupp]{axessibility}
\usepackage{times}
\usepackage{graphicx}
\usepackage{xspace}
\usepackage{comment}
\usepackage{soul}
\usepackage{pifont}
\usepackage{xparse}
\usepackage{floatpag}
\usepackage{amsmath,bm}
\usepackage{lipsum}

\usepackage{multirow}
\usepackage{rotating}
\usepackage{tabularx}
\usepackage{arydshln}
\usepackage{siunitx, booktabs}

\usepackage{dblfloatfix}
\usepackage{fontawesome}
\usepackage{array}

\newcommand{\algo}{\texttt{OneDiffusion}\xspace}

\usepackage[colorinlistoftodos, textwidth=30mm, shadow, textsize=small]{todonotes}
\definecolor{mypink}{rgb}{1, 0.4, 0.70}
\definecolor{babyblue}{rgb}{0.54, 0.81, 0.94}
\definecolor{citrine}{rgb}{0.89, 0.82, 0.04}
\definecolor{misocolor}{rgb}{0.16,0.27,0.86}
\definecolor{jbcolor}{rgb}{0.9,0.4,0.2}
\definecolor{bernacolor}{rgb}{0.9608,0.4863,0.00}
\definecolor{carlcolor}{rgb}{0.0,0.9863,0.30}
\definecolor{grey}{rgb}{0.3, 0.3, 0.3}
\definecolor{graphicbackground}{rgb}{0.96,0.96,0.8}
\definecolor{rouge1}{RGB}{226,0,38}  %
\definecolor{orange1}{RGB}{243,154,38}  %
\definecolor{jaune}{RGB}{254,205,27}  %
\definecolor{blanc}{RGB}{255,255,255} %
\definecolor{rouge2}{RGB}{230,68,57}  %
\definecolor{orange2}{RGB}{236,117,40}  %
\definecolor{taupe}{RGB}{134,113,127} %
\definecolor{gris}{RGB}{91,94,111} %
\definecolor{bleu1}{RGB}{38,109,131} %
\definecolor{bleu2}{RGB}{28,50,114} %
\definecolor{vert1}{RGB}{133,146,66} %
\definecolor{vert3}{RGB}{20,200,66} %
\definecolor{vert2}{RGB}{157,193,7} %
\definecolor{darkyellow}{RGB}{233,165,0}  %
\definecolor{lightgray}{rgb}{0.9,0.9,0.9}
\definecolor{darkgray}{rgb}{0.6,0.6,0.6}
\definecolor{babyblue}{rgb}{0.54, 0.81, 0.94}
\definecolor{citrine}{rgb}{0.89, 0.82, 0.04}
\definecolor{misogreen}{rgb}{0.25,0.6,0.0}
\definecolor{PalePurp}{rgb}{0.66,0.57,0.66}
\definecolor{todocolor}{rgb}{0.66,0.99,0.99}
\definecolor{pearOne}{HTML}{2C3E50}
\definecolor{pearTwo}{HTML}{A9CF54}
\definecolor{pearTwoT}{HTML}{C2895B}
\definecolor{pearThree}{HTML}{E74C3C}
\colorlet{titleTh}{pearOne}
\colorlet{bull}{pearTwo}
\definecolor{pearcomp}{HTML}{B97E29}
\definecolor{pearFour}{HTML}{588F27}
\definecolor{pearFith}{HTML}{ECF0F1}
\definecolor{pearDark}{HTML}{2980B9}
\definecolor{pearDarker}{HTML}{1D2DEC}
\definecolor{darkTurquoise}{HTML}{007C7D}

\usepackage[pagebackref,breaklinks,colorlinks,citecolor=cvprblue]{hyperref}

\def\paperID{13726} %
\def\confName{CVPR}
\def\confYear{2025}

\title{One Diffusion to Generate Them All}

\definecolor{cvprblue}{rgb}{0.21,0.49,0.74}
\definecolor{mydarkblue}{rgb}{0,0.08,1}
\definecolor{mydarkgreen}{rgb}{0.02,0.6,0.02}
\definecolor{mydarkred}{rgb}{0.8,0.02,0.02}
\definecolor{mydarkorange}{rgb}{0.40,0.2,0.02}
\definecolor{mypurple}{RGB}{111,0,255}
\definecolor{myred}{rgb}{1.0,0.0,0.0}
\definecolor{mygold}{rgb}{0.75,0.6,0.12}
\definecolor{myblue}{rgb}{0,0.2,0.8}
\definecolor{mydarkgray}{rgb}{0.66,0.66,0.66}

\author{Duong H. Le$^{1,*}$ \quad Tuan Pham$^{2,*}$ \quad Sangho Lee$^{1}$ \quad Christopher Clark$^{1}$ \\  
Aniruddha Kembhavi$^1$ \quad Stephan Mandt$^2$ \quad Ranjay Krishna$^{1,3}$ \quad Jiasen Lu$^1$ \\
\normalsize $^1$Allen Institute for AI \quad $^2$ University of California, Irvine \quad $^3$ University of Washington  \quad * Equal contribution\\
}

\begin{document}

\twocolumn[{
\renewcommand\twocolumn[1][]{#1}
\maketitle
\thispagestyle{empty}
\vspace*{-0.35in}
\centering
\captionsetup{type=figure}\includegraphics[width=\linewidth]{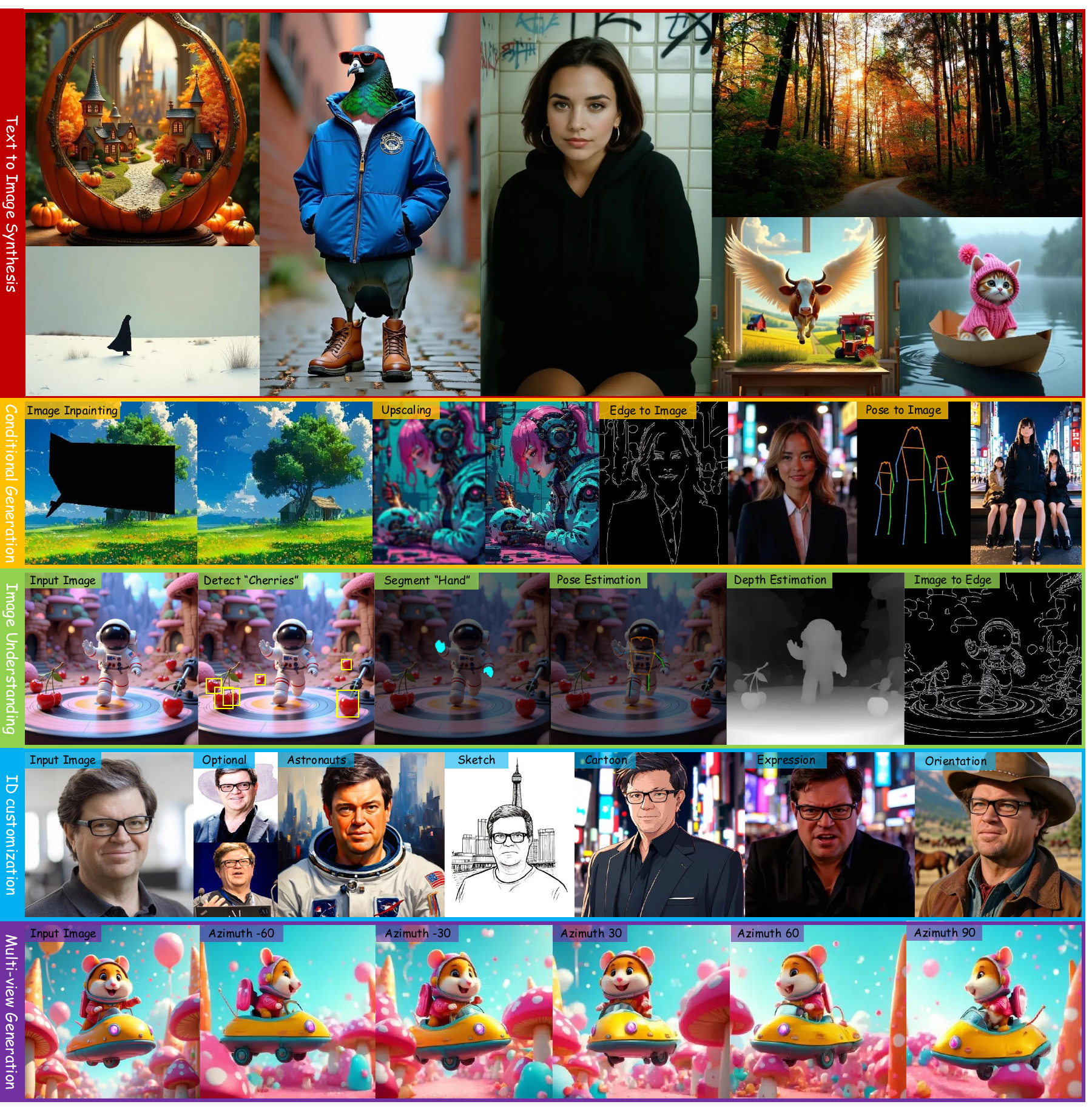}
\vspace{-20pt}
\captionof{figure}{
\algo is a unified diffusion model designed for both image synthesis and understanding across diverse tasks. 
It supports text-to-image generation (\textcolor{red}{red box}), conditional image generation from input images (\textcolor{orange}{orange box}) and it's reverse task Image understanding (\textcolor{green}{green box}). It can also perform ID customization (\textcolor{blue}{blue box}), and multi-view generation (\textcolor{purple}{purple box}) with arbitrary number of input and output images. 
}
\label{fig:teaser}
\vspace{-10pt}
}]

\maketitle
\thispagestyle{empty}
\clearpage

\begin{abstract}
We introduce \texttt{OneDiffusion}, a versatile, large-scale diffusion model that seamlessly supports bidirectional image synthesis and understanding across diverse tasks. 
It enables conditional generation from inputs such as text, depth, pose, layout, and semantic maps, while also handling tasks like image deblurring, upscaling, and reverse processes such as depth estimation and segmentation. Additionally, \texttt{OneDiffusion} allows for multi-view generation, camera pose estimation, and instant personalization using sequential image inputs.
Our model takes a straightforward yet effective approach by treating all tasks as frame sequences with varying noise scales during training, 
allowing any frame to act as a conditioning image at inference time. 
Our unified training framework removes the need for specialized architectures, supports scalable multi-task training, and adapts smoothly to any resolution, enhancing both generalization and scalability.
Experimental results demonstrate competitive performance across tasks in both generation and prediction such as text-to-image, multiview generation, ID preservation, depth estimation and camera pose estimation despite a relatively small training dataset.
Our code and checkpoint are freely available at \url{https://github.com/lehduong/OneDiffusion}.

\end{abstract}
    
\section{Introduction}
\label{sec:introduction}

Diffusion models, particularly in text-to-image (T2I) generation, have recently achieved remarkable results. Models such as DALL-E \citep{ramesh2021zero}, Imagen \citep{ramesh2021zero}, and Stable Diffusion \citep{rombach2022high, podell2023sdxl, esser2024scaling} have established new benchmarks for generating high-quality, photorealistic images from text prompts. Additionally, recent studies have demonstrated the effectiveness of diffusion models in various other computer vision tasks, such as depth estimation~\cite{ke2024repurposing} or optical flow estimation \citep{saxena2024surprising, luo2024flowdiffuser}, \etc. However, despite these advancements, diffusion models are typically trained individually for either T2I generation or specific tasks.   

In contrast, large language models (LLMs) (\eg GPT-4~\citep{achiam2023gpt}) have demonstrated their ability to function as universal models. They can perform a wide range of tasks across different domains without the need for task-specific modules, and can effectively handle tasks they have not been explicitly trained in a \textit{zero-shot} manner. 
This universality has been immensely valuable; it has dramatically simplified using training and scaling these models, and ultimately led to better performance. This incentivizes us to ask whether diffusion models can become universal in a similar way.

Designing a unified architecture for diverse image synthesis tasks presents significant challenges.
Current methods often depend on external add-ons to handle new tasks. For example, ControlNet \citep{zhang2023adding} or T2I-Adapter \citep{mou2024t2i} require specialized modules to encode the conditional inputs, and personalization models typically require encoding the identity through a pretrained facial recognition network and adding auxiliary losses to preserve identity \cite{ye2023ip, wang2024instantid, guo2024pulid}. 
Additionally, tasks vary widely in their input requirements. For instance, multi-view generation alone requires handling arbitrary input-output view combinations, posed or unposed images, and camera pose conditioning \citep{gao2024cat3d, shi2023mvdream, wang2023imagedream, liu2023syncdreamer, kong2024eschernet}, while image understanding tasks require diverse outputs such as depth, pose, or segmentation.
Finally, existing training recipes are often tightly tuned to particular tasks and therefore cannot be relied on to generalize between tasks.

In this work, we present \algo~-- a unified diffusion model that seamlessly supports bidirectional image synthesis and understanding across diverse tasks. 
Our approach enables a single model to perform multiple tasks without the need for external losses and add-ons.
Inspired by recent advances in diffusion models for sequential data \citep{zhang2024tedi, ruhe2024rolling, chen2024diffusion}, we model all conditions and target images as a sequence of ``views'' with varying noise levels during training.
At inference, any of the views can be used as a conditional input, or set to noise and then used to generate an output image. Conditioning text can also be changed to define the task, and specify additional conditioning details (\eg camera pose). The simple, but flexible, framework allows our model to support many kinds image generation and image understanding tasks with a unified architecture and training objective.

To demonstrate how general purpose our training algorithm is, we train \algo completely from scratch. First, we train on text-to-image task to equip the model with general image synthesis abilities, then on our One-Gen dataset to learn the full set of tasks. Our final model has 2.8 billion parameters and is equipped with a diverse set of skills, shown in Figure~\ref{fig:teaser}. The model also adapts naturally to various resolutions, enabling zero-shot high-resolution generation even when such resolutions were not encountered during training. 

We evaluate \algo on a diverse set of both generative and predictive tasks.  On T2I, \algo efficiently generates high-quality images while utilizing fewer number of parameters. 
In the multiview generation task, \algo demonstrates performance comparable to state-of-the-art methods that are specifically designed and exclusively trained for this purpose.
We also show that \algo supports novel conditioning setups, such as text-to-multi-view and image-to-multi-view. For high-variability tasks like face identification from a single image, the model is capable of generating multiple consistent images featuring diverse expressions and poses, demonstrating strong generalization to unseen domains.

\section{Related work}
\paragraph{Diffusion models for generative tasks} 
Recent advancements in diffusion models have greatly improved image generation capabilities, with models like Stable Diffusion \cite{rombach2022high, podell2023sdxl, esser2024scaling, betker2023improving, baldridge2024imagen, zhuo2024lumina} setting new standards in text-to-image synthesis. Beyond general image generation, controllable diffusion models such as ControlNet \citep{zhang2023adding} and T2I-Adapter \citep{mou2024t2i} allow fine-grained control via auxiliary inputs like edge or depth maps. Similar structured conditioning has been applied to inverse problems \cite{song2023pseudoinverse,pandey2024fast,pandey2025variational}, enabling applications such as super-resolution or inpainting. Meanwhile, instruct-Pix2Pix \citep{brooks2023instructpix2pix}  introduces natural language-guided image editing, making these tools more user-friendly.
For personalized applications, identity-focused models, including IP-Adapter \citep{ye2023ip}, InstantID \citep{wang2024instantid}, PhotoMaker \citep{li2024photomaker}, and PuLiD \citep{guo2024pulid}, personalize generation by conditioning on reference images. 
Moreover, in multi-view generation, recent methods  \cite{shi2023mvdream, wang2023imagedream, liu2023syncdreamer, gao2024cat3d}, 
employ camera ray embeddings or 3D geometry to achieve consistent viewpoints. Together, these innovations showcase the versatility of diffusion models in delivering controllable, personalized, and multi-perspective image synthesis.

\vspace{-5mm}
\paragraph{Diffusion models for predictive tasks} Beyond image generation and manipulation, diffusion models have also proven effective for predictive tasks within computer vision. Marigold \citep{ke2024repurposing} fine-tunes the Stable Diffusion model \citep{rombach2022high} to perform monocular depth estimation, demonstrating the adaptability of diffusion models for prediction-based applications. Furthermore, diffusion models have been utilized for optical flow estimation, as shown in the works of Saxena et al. \citep{saxena2024surprising} and Luo et al. \citep{luo2024flowdiffuser}, where the models predict pixel-level motion between consecutive frames. Additionally, Li et al. \citep{li2023open} trained a diffusion model for open-vocabulary semantic segmentation, showcasing the potential of these models for more complex vision tasks. Prior works have attempt to unify diffusion model for predictive tasks \cite{geng2024instructdiffusion, gan2023instructcv}. These studies show that diffusion models are not only useful for generating images but also highly effective for various predictive tasks in computer vision.

\vspace{-5mm}
\paragraph{Unified diffusion models} Several attempts have been made to unify diffusion model for different type of controls \citep{zhao2023uni, qin2023unicontrol, xu2023versatile}. However, they are limited to utilization of multiple image conditions. These models usually requires to design complicated adapters for different conditions. \citep{lu2022unified, lu2023uio2, team2024chameleon, zhou2024transfusion} propose unified models for language and images. 
Concurrently, \cite{xiao2024omnigen} propose finetuning multimodal large language model with diffusion objective on diverse tasks like text-to-image, editing, and subject-driven generation etc. In contrast, our model distinguishes itself by leveraging bidirectional capabilities of diffusion models and addressing a wide range of diverse tasks.

\begin{figure*}
    \centering
    \includegraphics[width=\linewidth]{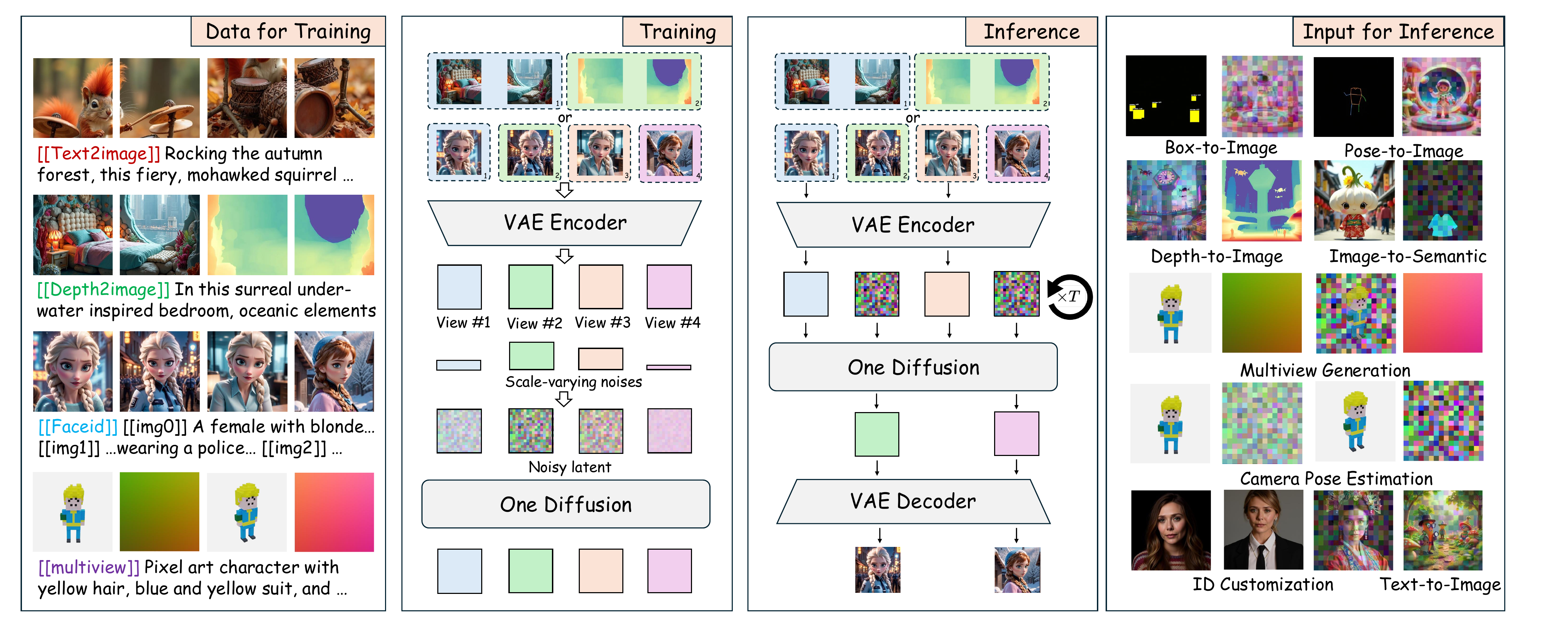}
    \vspace{-15pt}
    \caption{Illustration of training and inference pipeline for \algo. We encode the desired task for each sample via a special task token. During training we independently sample different diffusion timesteps for each view and add noise to them accordingly. In inference, we replace input image(s) with Gaussian noises while setting timesteps of conditions to $0$.}
    \label{fig:pipeline}
    \vspace{-3pt}
\end{figure*}

\section{Methodology}
\label{sec:method}
\subsection{Flow matching for generative modeling}
Flow matching~\cite{lipman2023flow, liu2022flow, albergo2022building} is a framework for training continuous-time generative models by learning a time-dependent vector field that transports between two probability distributions. More specifically, a time-dependent vector field $u_t: [0, 1] \times \mathbb{R}^d \rightarrow \mathbb{R}^d$ governs the transformation from a base distribution $p_0$ to the target distribution $p_1 \approx q$  through an ODE $dx = u_t(x) dt$. 

The solution of this ODE is a flow $\phi_t: [0, 1] \times \mathbb{R}^d \rightarrow \mathbb{R}^d$ with initial condition $\phi_0(x) = x$, and this flow characterizes a push-forward operation $p_t = [\phi_t]_\# p_0$, in which $p_t$ is the density of samples $x \sim p_0$ transported by $u$ from time $0$ to time $t$. The goal is approximate this ODE using a learned time-dependent vector field parameterized as a neural network $v_\theta (t, x)$. Due to the intractable nature of $u_t$, \cite{lipman2023flow} proposed to learn $v_\theta (t, x)$ using the conditional flow matching (CFM) objective:
\begin{equation}
    \mathcal{L}_{\mathrm{CFM}}(\theta):=\mathbb{E}_{t, q(z), p_t(x \mid z)}\left\|v_\theta(t, x)-u_t(x | z)\right\|^2
\end{equation}
This objective is equivalent to the original flow matching objective, and only requires the samples from the target distribution and a suitable conditional probability path.

\subsection{Proposed Approach}
\paragraph{Objective} We cast the problem of image generation with multimodal conditions as sequential modeling. Inspired by previous work on diffusion model for sequential data \cite{zhang2024tedi, ruhe2024rolling, chen2024diffusion}, we jointly model all conditions and target images as a sequence of ``\emph{views}''. Note that the number of views $N$ is determined by tasks. Particularly, $N=1$ for text-to-image tasks, $N=2$ for image-to-image translation such depth/pose/image editing, etc, $N>2$ for multiview generation or ID customization.

Mathematically, let $N$ views $\{\mathbf{x}_i\}_{i=1}^N \in \mathbb{R}^{H \times W \times D}$ be sampled from a training dataset $q(\mathbf{x}_1, ..., \mathbf{x}_N)$. 
Given time variables $t_i$, our goal is to learn a function $v_\theta(t_1, ..., t_N, \mathbf{x}_1, ..., \mathbf{x}_N): [0, 1]^N \times \mathbb{R}^{N \times H \times W \times D} \rightarrow \mathbb{R}^{H \times W \times N \times D}$. Intuitively, $v_\theta$ serves as a generalized time-dependent vector field where each input $\mathbf{x}_i$ paired with its respective time variable $t_i$.

Learning $v_\theta$ enables arbitrary conditional generation, where any subset of views can be selected as conditions to generate the remaining views, as explained below. This setup allows us to dynamically configure the generation process, supporting flexible applications across a range of generative tasks.
\vspace{-2mm}

\paragraph{Training} Our training pipeline is visualized on the left side of Figure \ref{fig:pipeline}. 
At each training step, we independently sample $t_i \sim \text{LogNorm}(0, 1)$ \cite{esser2024scaling} and Gaussian noise $\epsilon_i \sim \mathcal{N}(0, I)$. 
This results in different noise levels for each views. We apply an interpolation-based forward process:
\begin{equation}
    \mathbf{x}_i^{t_i} = \alpha_{t_i} \mathbf{x}_i + \beta_{t_i} \epsilon_i
\end{equation}
where $\alpha_t$ and $\beta_t$ satisfy the boundary conditions $\alpha_0 = 0, \alpha_1 = 1$ and $\beta_0 = 1, \beta_1=0$. Similar to \cite{zhuo2024lumina}, we adopt the linear interpolation schedule:
\begin{equation}
    \mathbf{x}_i^{t_i} = t_i \mathbf{x}_i + (1 - t_i) \epsilon_i
\end{equation}
the corresponding velocity field $u_i$
for each view $\mathbf{x}_i$ is:
\begin{equation}
    u_i(t_i, \mathbf{x}_i) = \mathbf{x}_i - \epsilon_i
\end{equation}
with the aggregated target as $u=(\mathbf{x}_1 - \epsilon_1, ..., \mathbf{x}_N - \epsilon_N) \in \mathbb{R}^{N \times H \times W \times D}$, our training loss is the joint flow-matching objective:
\begin{equation}
    \mathcal{L}(\theta) = \mathbb{E} \left[ \| v_\theta(t_1, ..., t_N, \mathbf{x}_1, ..., \mathbf{x}_N) - u\|^2 \right]
    \label{eq:loss}
\end{equation}
This flow matching objective~\cite{lipman2023flow, liu2022flow, albergo2022building} guides the model to learn the optimal velocity field $v_\theta$ by minimizing the difference from the target velocity field $u$. 

\paragraph{Inference} 

Our framework allows for both joint sampling and conditional sampling across any chosen set of views. In details, we define the target views we want to sample as  $\mathbf{x}_K = \{\mathbf{x}_i\}_{i \in K}$, and the set of conditional views as $\mathbf{x}_{\setminus K} = \{\mathbf{x}_i\}_{i \notin K}$. To perform conditional sampling, we start by initializing the target views $\mathbf{x}_K$ as Gaussian noise. At each timestep $t$, we compute the corresponding time-dependent vector field $v_\theta^K(t, \mathbf{x} | \bar{\mathbf{x}}_{\setminus K})$ by fixing the conditional views to their known values $\mathbf{x}_{\setminus K} = \bar{\mathbf{x}}$ and setting their time variables to zero $t_{\setminus K} = 0$; while keeping the time variables of the target views as $t_K = t$:
\begin{equation}
    v_\theta^K(t, \mathbf{x} | \mathbf{x}_{\setminus K} = \bar{\mathbf{x}}) = v_\theta(t_K=t, t_{\setminus K}=0, \mathbf{x}_K=\mathbf{x},  \mathbf{x}_{\setminus K}=\bar{\mathbf{x}})
\end{equation}
Note that unlike $v_\theta$, $v_\theta^K$ is a valid time dependent vector field, as all the views in $K$ now has the same $t$. Thus, by integrating this vector field using an ordinary differential equation (ODE) solver, we can generate the conditional samples we are interested in. We illustration the inference on the right side of Figure \ref{fig:pipeline}. 

\begin{figure*}[!h]
    \centering
    \includegraphics[width=\linewidth]{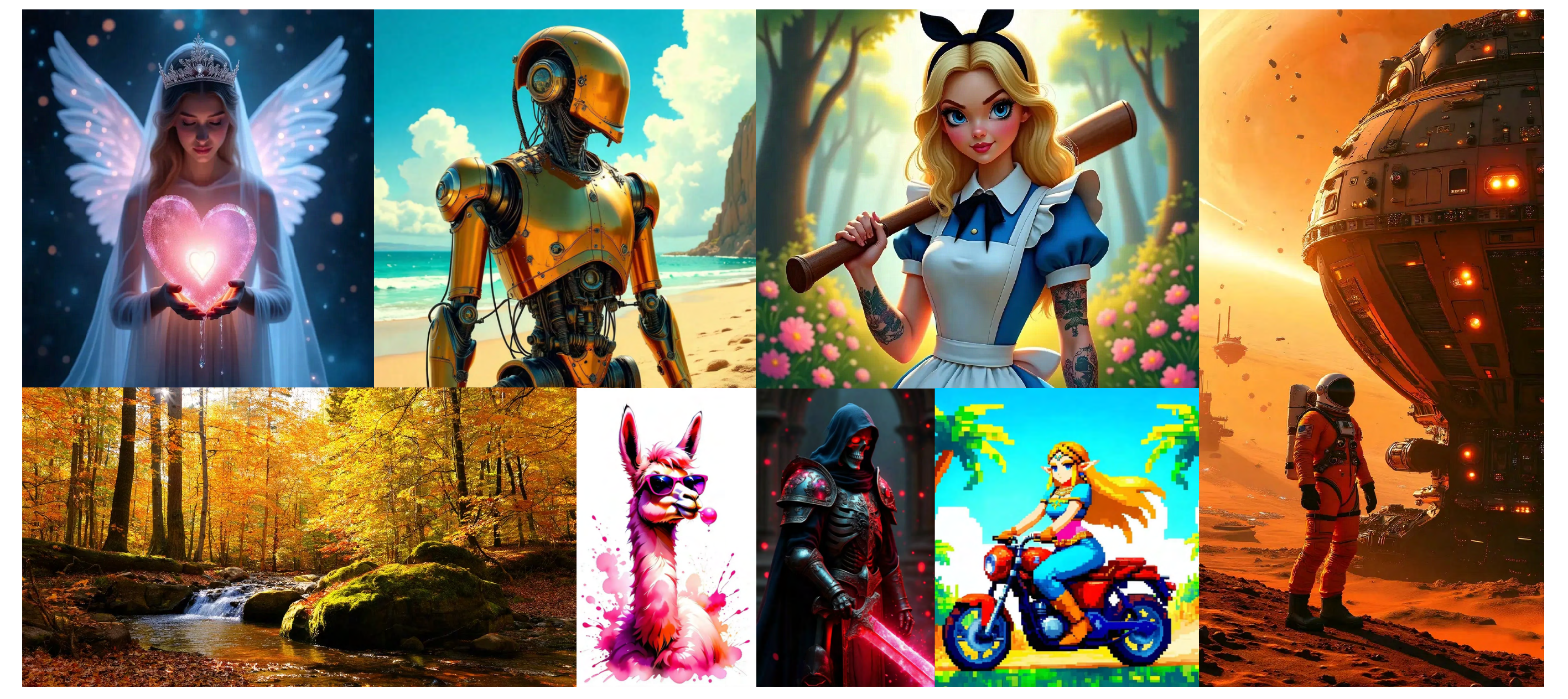}
    \vspace{-15pt}
    \caption{High-resolution samples \textbf{from text} of our \algo model, showcasing its capabilities in precise prompt adherence, attention to fine details, and high image quality across a wide variety of styles.}
    \label{fig:demo_text2image}
    \vspace{-4pt}
\end{figure*}
\vspace{-0.25mm}

\subsection{Implementation Details}
\paragraph{Model architecture} We adopt the Next-DiT architecture \cite{zhuo2024lumina} in our model. By leveraging a full transformer-based architecture, our model can work with different numbers of views $N$. We independently encode each frame \ie images and conditions as latent $\mathbf{z} \in \mathbb{R}^{N\times H \times W \times C}$ with a VAE tokenizer \cite{esser2024scaling} and concatenate them in $N$ dimension. With flexible $N$, our approach establishes a universal framework which supports diverse input-modality with variable length. Following \cite{zhuo2024lumina}, we also apply 3D RoPE \cite{su2024roformer} for positional encoding, enabling generalization to different resolutions and aspect ratios.
\vspace{-2mm}

\paragraph{Text-to-Image (1 views)} With only a single ``view'', training and inference follow the same process as standard text-to-image diffusion models. We prepend the task label \texttt{[[text2image]]} to the caption to specify the task.
\vspace{-2mm}

\paragraph{Image-to-Image (2 views)} We set the first view as the target image and the second as the conditioning input. During inference, we can use one or both views for generation, and the model is trained to produce the target image. For tasks like bounding box or semantic map generation, we add the hexadecimal color code and class label to the prompt. For instance, to segment a mouse with a yellow mask, the prompt is:
\texttt{[[semantic2image]] \textless \#FFFF00 yellow mask: mouse\textgreater\space photo of a \ldots } Further details are provided in the appendix. 
\vspace{-2mm}

\vspace{-1mm}
\paragraph{ID Customization (2-4 views)}  We sample images of the same individual across views, concatenating captions for each input image and using a token \texttt{[[imgX]]} to denote each image. We also prepend the task label \texttt{[[faceid]]} to the captions.
At inference, we can condition on an arbitrary number of images and generate multiple outputs, leading to more consistent results.

\vspace{-3mm}
\paragraph{Multiview Generation (4-12 views)}
Inspired by \cite{gao2024cat3d}, we use Plücker ray embeddings to represent camera poses. 
For each image patch, we calculate Plücker coordinates as $\bm{r} = (\bm{o} \times \bm{d}, \bm{d})$ using its ray origin $\bm{o}$ and direction $\bm{d}$. 
The result embedding has dimensions $H/8 \times W/8 \times 6$, matching the spatial size of the latent,  and is replicated across channels to form a $16$ channel embedding. Unlike \cite{gao2024cat3d}, we treat ray embedding as a independent ``view'' following image latents as a unified sequence rather than concatenating by channels. This design allows flexible denoising, enabling multi-view image generation conditioned on camera poses or sampling ray embeddings to predict poses from image conditions, similar to RayDiffusion \cite{zhang2024cameras}. We scale the ray embeddings to have unit variance, as in \cite{rombach2022high}.

\begin{figure*}
    \centering
\includegraphics[width=\linewidth]{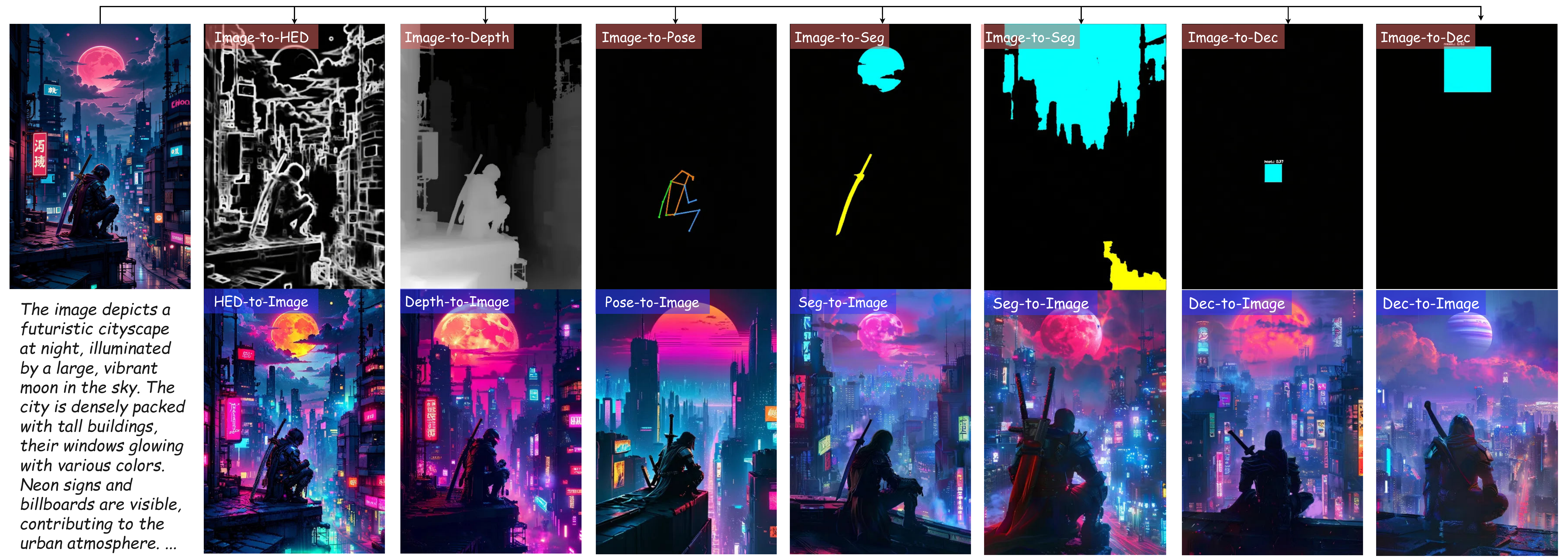}
    \vspace{-15pt}
    \caption{Illustration of our model capability to generate HED, depth, human pose, semantic mask, and bounding box from input image. For semantic segmentation, we segment the {\color{darkyellow} sword} (highlighted in yellow) and the {\color{cyan}moon} (highlighted in cyan) the first example, while segmenting {\color{darkyellow} road} (yellow), {\color{cyan}sky} (cyan) in the second. For object detection, We localize the {\color{cyan}head} and {\color{cyan}moon} (both highlighted in cyan). Leveraging these conditions, we can reverse the process to recreate a variant of the input image based on the same caption. Additionally, we can edit the image by modifying specific elements, such as replacing the moon with Saturn (last example). }
    \label{figure:bidrectional_image_generation}
    \vspace{-4pt}
\end{figure*}

As with other tasks, we prepend the task label \texttt{[[multiview]]} to the caption. During inference, we can substitute images or Plücker ray embeddings with Gaussian noise for multi-view generation and camera pose estimation, respectively.

\vspace{-3mm}
\paragraph{Training details} 
Our model is trained from \emph{scratch} using a flow-matching objective. Similar to prior works \cite{esser2024scaling, pixart}, we use a three stage training recipe.
In the \textbf{first stage}, we pretrained the text-to-image model with resolution of $256^2$ (500K steps) and $512^2$ (500K steps). In the \textbf{second stage}, we continue training on a mixed of tasks, using $512^2$ for T2I and $256^2$ for other tasks, for a total of 1M steps. 
Finally, in the \textbf{last stage}, we finetune the model at a high resolution of ($1024$) for T2I. 
For ID customization fine-tuning, we use 2-5 views. For fewer views (2-3), we apply a resolution of $512^2$, while for more views, we use $256^2$ resolution.

During training, we use an in-batch sampling strategy at each stage, sampling tasks (T2I, Image-to-Image, ID customization, and multiview generation) with equal probability. The noise scheduler's shift value is set to 3, as suggested in \cite{esser2024scaling}. We use AdamW optimizer with learning rate $\eta=0.0005$. Training is performed on a TPU v3-256 pod with a global batch size of 256 in the first two phases, and the final fine-tuning stage is completed on 64 H100 GPUs using the same configuration. 

\section{One-Gen Datasets}
\label{sec:experiment_dataset}
\paragraph{Text-to-Image} We leverage both public and internal (synthetic) datasets. The public datasets including: PixelProse \citep{singla2024pixels}, Unsplash,
Coyo \citep{kakaobrain2022coyo-700m}, JourneyDB \cite{pan2023journeydb}. Additionally, we use a 10M internal synthetic dataset consisting of images re-captioned with LLaVA-NeXT \citep{liu2024llavanext} and Molmo \citep{deitke2024molmo}. The length of the text description for each image varies from $100$ to $150$ words. When an original prompt is available, we use both the LLM-generated caption and the original caption.
\vspace{-5mm}

\paragraph{Image-to-Image} For simpler tasks \eg deblurring, inpainting, image generation from canny edge, or upscaling, we use a $1M$-sample subset of our synthetic data and apply the corresponding pre-processor for each image to create an input condition. For more complex tasks, we create a synthetic dataset from outputs generated by Midjourney, Stable Diffusion, and Flux-dev following the below process:
\begin{itemize}
    \item \textbf{Semantic Map and Detection} For each image, we use the LLaVA-NeXT \citep{liu2024llavanext} model to identify entities or subjects (e.g., person, shirt, dog, building), with a maximum of 10 entities per image. Based on these subject names from LLaVA-Next, we perform semantic segmentation using SAM \citep{kirillov2023segment} and extract bounding boxes. Each class is assigned a random color from a predefined list. This dataset contains 350K triplets consisting of a semantic map, bounding box, and the original image.
    \item \textbf{Depth Map} We generate the depth dataset by applying DepthAnything-v2 \citep{yang2024depth} to 500K images sampled from various datasets, including both real and synthetic images. Additionally, we caption 40K images from Hypersim dataset \cite{roberts2021hypersim} with LLaVA-NeXT and incorporate these into the training set.
    \item \textbf{Human Poses} We collect a different subset with 50K images, primarily featuring human for pose conditioning. We use YOLOv5 to detect the bounding boxes for region of interests and apply ViTPose \citep{xu2022vitpose} for pose estimation.
\end{itemize}
\vspace{-4mm}
\paragraph{ID Customization} We collect a dataset of celebrities and characters from games and movies by from publicly available images. After filtering to ensure each subject has at least four images and removing NSFW content, the dataset includes approximately 60K subjects and a total of 1.3M images. We caption these images using the LLaVA-NeXT.
\vspace{-4mm}
\paragraph{Multiview Generation} We use the DL3DV-10K dataset \citep{ling2024dl3dv}, Objaverse \citep{deitke2022objaverse}, CO3D \cite{reizenstein2021common}. For Objaverse dataset, we utilize the 80K filtered split from LGM \citep{tang2024lgm} and caption provided by Cap3D \citep{luo2024scalable}. In the DL3DV dataset, we sample an image from each scene and caption it using LLaVA-Next. For CO3D, we exclude captions and include only the task token in the text input.
\vspace{-2mm}

\begin{figure*}
    \centering
    \includegraphics[width=\linewidth]{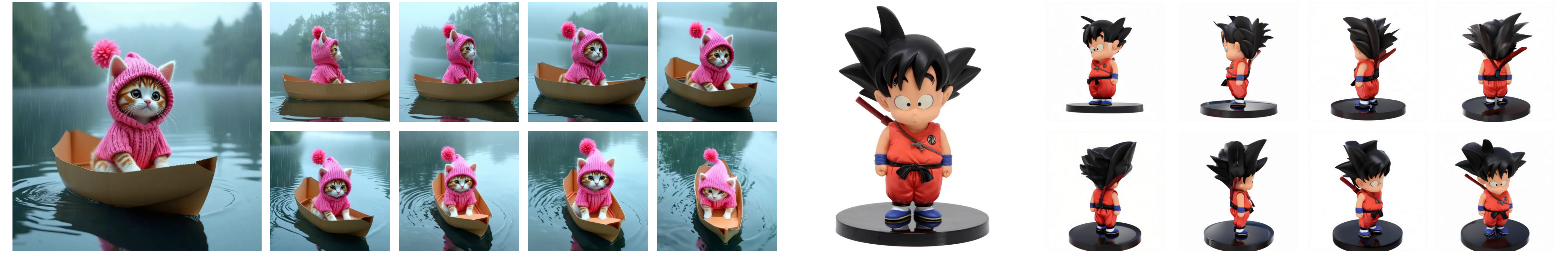}
    \vspace{-10pt}
    \caption{Illustration of the multiview generation with single input image. We equally slice the azimuth in range of $[-45, 60]$ and elevation in range of $[-15, 45]$ for the left scenes. For the right scene, the azimuth range is set to $[0; 360]$ and elevation range is set to $[-15; 15]$.}
    \label{fig:multiview_cat}
    \vspace{-6pt}
\end{figure*}

\section{Experiments}
We evaluate our \algo model on broad range of image generation and understanding tasks. We do not perform task-specific finetuning in any results. Details about additional qualitive examples are in the Appendix.

\subsection{Text-to-Image}

Qualitative results of \algo for text-to-image task is illustrated in Figure \ref{fig:demo_text2image}. Thanks to the diversity of our One-Gen dataset, the model can handle various art styles, spanning both artistic and photorealistic designs. 

Following previous works \citep{esser2024scaling}, we evaluated the text-to-image capabilities of our model on GenEval benchmark \citep{ghosh2024geneval}. For each prompt, we generate 4 images using Euler solver with $100$ steps and guidance scale of $5$. The results for \algo, along with those of baseline models, are presented in Table \ref{tab:geneval}. Our model demonstrates strong performance compared to similarly sized baselines, excelling in multitasking capabilities despite being trained on a relatively smaller dataset.This performance is largely attributed to the diversity of the dataset and the comprehensive captions provided for each sample.
\begin{table}
\resizebox{\linewidth}{!}{%
\begin{tabular}{lccc}
\hline
Methods & Params (B) & \# Data (M) & GenEval $\uparrow$ \\
\hline
LUMINA-Next \cite{zhuo2024lumina}     & 2.0  & 14     & 0.46 \\
PixArt-$\Sigma$ \cite{pixart_sigma}   & 0.6  & 33     & 0.54 \\
SDXL \cite{podell2023sdxl}            & 2.6  & --     & 0.55 \\
PlayGroundv2.5 \cite{playground}      & 2.6  & --     & 0.56 \\
IF-XL                                 & 5.5  & 1200   & 0.61 \\
SD3-medium \cite{esser2024scaling}    & 2.0  & 1000   & 0.62 \\
Hunyuan-DiT \cite{hunyuandit}         & 1.5  & --     & 0.63 \\
DALLE3                                & --   & --     & \underline{0.67} \\
FLUX-dev                              & 12.0 & --     & \underline{0.67} \\
FLUX-schnell                          & 12.0 & --     & \textbf{0.71} \\
\hline
\algo                                 & 2.8  & 75     & \emph{0.65} \\
\hline
\end{tabular}%
}
\vspace{-4pt}
\caption{Comparison of text-to-image performance on the GenEval benchmark at a resolution of 1024 $\times$ 1024.}
\vspace{-4pt}
\label{tab:geneval}
\end{table}

\subsection{Controllable Image generation}
We show the experiment with image-to-image translation using various source domains, including HED, depth map, human pose, semantic map, bounding boxes. We report the qualitative results in Figure \ref{figure:bidrectional_image_generation} and \ref{figure:appendix_bidirectional_examples} in appendix. Generated images of \algo consistently conform various types of conditions by purely utilizing attention mechanisms and supplementary information from captions.

\subsection{Multiview Generation}

\begin{figure*}
    \centering
    \includegraphics[width=\linewidth]{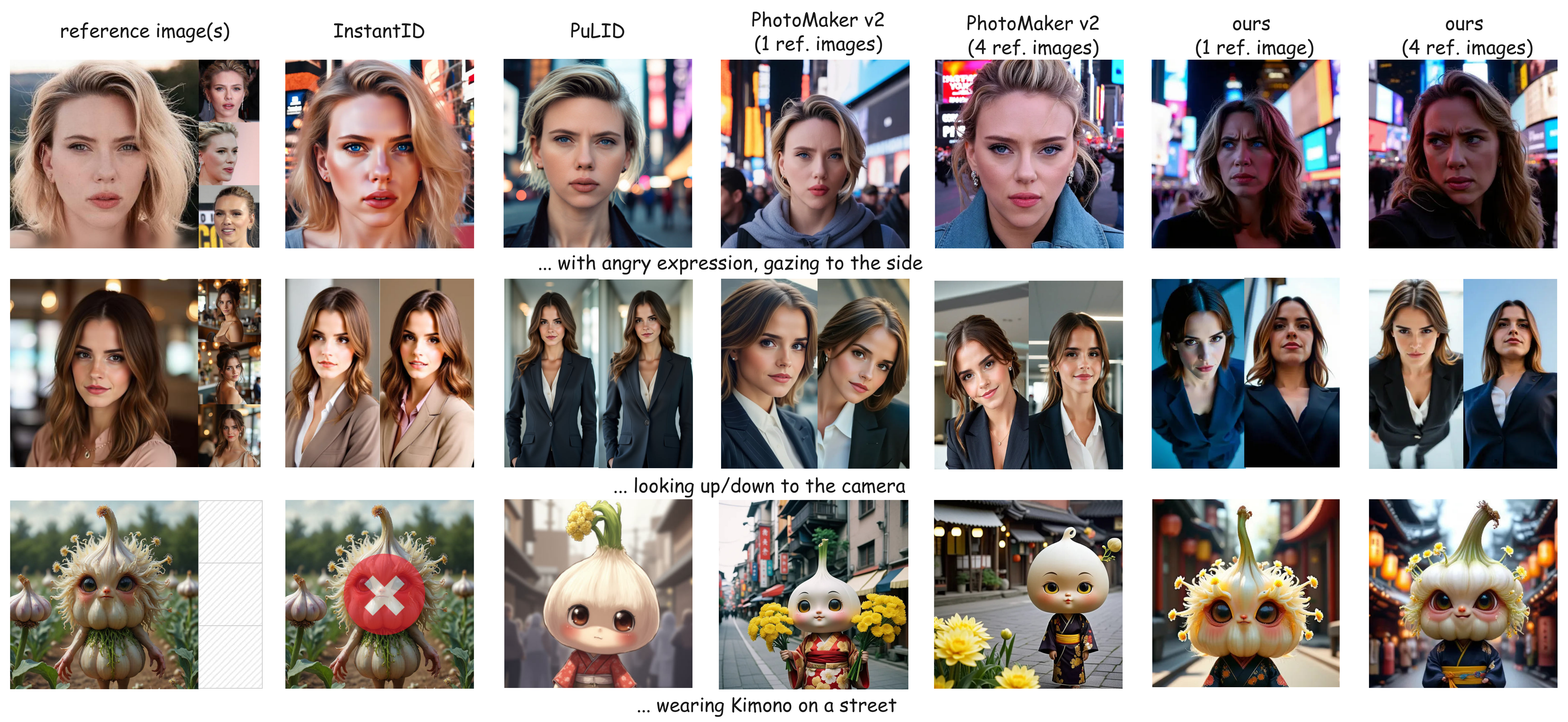}
    \caption{
    Illustration of ID customization using reference images. Unlike prior methods that rely on face embeddings and often fail to generalize, our model demonstrates superior generalization. It effectively adjusts facial expressions and gaze directions (first row), changes viewpoints (second row), and even customizes non-human IDs (third row). All results in the third row are generated from a single reference image, while InstantID fails as its face detector cannot detect faces in the input. 
    }
    \label{fig:faceid_expression_comparison}
    \vspace{-4pt}
\end{figure*}

\begin{table}[t]
\centering
\resizebox{0.85\linewidth}{!}{%
\begin{tabular}{lccc}
\hline
Model & Condition & PSNR $\uparrow$ \\
\hline
Zero123 \citep{liu2023zero} & 1-view & 18.51 \\
Zero123-XL \citep{deitke2024objaverse} & 1-view & 18.93 \\
\hline
& 1-view & 20.24 \\
EscherNet \citep{kong2024eschernet} & 2-view & 22.91 \\
& 3-view & 24.09 \\
\hline
& 1-view & 19.01 \\
& 2-view (unknown poses) & 19.83 \\
\algo & 2-view (known poses) & 20.22 \\
& 3-view (unknown poses) & 20.64 \\
& 3-view (known poses) & 21.79 \\
\hline
\end{tabular}%
}
\vspace{-4pt}
\caption{Comparison of NVS metrics across different number of condition view settings. Increasing the number of condition views improves the reconstruction quality.}
\vspace{-4pt}
\label{table:mvquant}
\end{table}

We assess our method’s multiview generation capabilities using the Google Scanned Object dataset. Table \ref{table:mvquant} compares our approach (\algo) with state-of-the-art methods like Zero123, Zero123-XL, and EscherNet, which are tailored for multiview tasks. Unlike these, \algo supports variable conditional inputs and can handle additional views with unknown camera poses due to its flexible denoising framework.

In Table \ref{table:mvquant}, \algo outperforms Zero123 and Zero123-XL in the 1-view condition and maintains strong results with unknown poses, e.g., a PSNR of 19.83 (2-view, unknown) vs. 20.22 (known), and 20.64 (3-view, unknown) vs. 21.79 (known). Figure \ref{fig:multiview_cat} shows consistent multiview outputs from a single front-view image, with more examples in Appendix Figures \ref{fig:appendix_multiview_demo_1} and \ref{fig:appendix_multiview_demo_2}. Our framework also enables text-to-multiview generation using only camera poses, as shown in Figure \ref{fig:appendix_text_to_multiview}.

\subsection{ID Customization}

We further evaluate \algo on ID customization tasks, which involve using one or multiple ID images as inputs for personalized generation.
To assess performance, we compare with STOA methods, including InstantID \cite{wang2024instantid}, PuLID \cite{guo2024pulid}, and PhotoMaker \cite{li2024photomaker}, using both qualitative and quantitative analyses. Our evaluation extends beyond the standard benchmark (Unsplash-50 \cite{gal2024lcm}) to test generalization on ID customization tasks, such as varying expressions, viewpoints, and even non-human images.

Figure \ref{fig:faceid_expression_comparison} illustrates examples of altering facial expressions and gaze directions (first row), changing viewpoints (second row), and customizing non-human IDs (third row). 
Our method achieves success in these tasks, where all other methods fail.
Unlike previous approaches that rely on face embeddings and primarily ``replicate'' the original face, \algo employs attention mechanisms between images and text conditions. This enables flexible end-to-end training and generates more expressive outputs, making our method suitable for a wider range of applications. Intuitively, the mechanism that ensures consistent multiview generation also proves effective for manipulating camera angles in ID customization, highlighting its adaptability across related applications. Additional visualizations are provided in Figure \ref{fig:appendix_faceid_demo_1} and \ref{fig:appendix_faceid_demo_2}.

\begin{table}[t]
\centering
\resizebox{0.6\linewidth}{!}{%
\begin{tabular}{lcc}
\toprule
Method & ID $\uparrow$ & CLIP-T $\uparrow$ \\
\midrule
PhotoMaker \cite{li2024photomaker} & 0.193 & 27.38 \\
InstantID \cite{wang2024instantid} & 0.648 & 26.41 \\
PuLID \cite{guo2024pulid} & \textbf{0.654} & \textbf{31.23} \\
Ours & 0.283 & 26.80 \\
\bottomrule
\end{tabular}%
}
\vspace{-4pt}
\caption{Quantitative results on Unsplash-50.}%
\vspace{-6pt}
\label{tab:unsplash_50}
\end{table}

We also present the quantitative results on the Unsplash-50 \cite{gal2024lcm} benchmark in Table \ref{tab:unsplash_50}. This benchmark focuses solely on style changes and re-contextualization, where PuLID \cite{guo2024pulid} demonstrates strong performance by leveraging embeddings from ID encoder networks trained on human faces for discrimination tasks. While this approach effectively preserves the identity traits of input images, it faces significant limitations when handling more complex face manipulations.

\subsection{Depth Estimation}

\begin{table}
\resizebox{\linewidth}{!}{%
\begin{tabular}{lcccccc}
\hline
\multirow{2}{*}{Method} & \multicolumn{2}{c}{NYUv2} & \multicolumn{2}{c}{DIODE} \\
 & AbsRel$\downarrow$ & $\delta_1\uparrow$ & AbsRel$\downarrow$ & $\delta_1\uparrow$ \\
\hline
DiverseDepth \citep{yin2020diversedepth}   & 11.7 & 87.5 & 37.6 & 63.1 \\
MiDaS  \citep{ranftl2020towards}         & 11.1 & 88.5 & 33.2 & 71.5 \\
DPT \citep{ranftl2020towards}            & 9.8  & 90.3 & \textbf{18.2} & 75.8 \\
LeReS  \citep{yin2021learning}         & 9.0  & 91.6 & 27.1 & 76.6 \\
Omnidata \citep{eftekhar2021omnidata}       & 7.4  & 94.5 & 33.9 & 74.2 \\
HDN  \citep{zhang2022hierarchical}           & 6.9  & 94.8 & 24.6 & \textbf{78.0} \\
Marigold \citep{ke2024repurposing}       & \underline{6.0}  & \underline{95.9} & 31.0 & 77.2 \\
DepthAnything-2 \citep{yang2024depth} & \textbf{4.6}  & \textbf{97.7} & 27.1 & 74.8 \\
\hline
Ours            & 6.8 & 95.2 & 29.4 & 75.2 \\
\hline
\end{tabular}%
}
\caption{Comparison of depth estimation methods on NYUv2 and DIODE datasets. \algo achieves competitive performance compared to previous depth estimation methods.}
\label{tab:depth_comparison}
\vspace{-6pt}
\end{table}

For image understanding tasks, we evaluate our model's performance on monocular depth estimation using standard benchmarks: NYUv2 \cite{silberman2012indoor} and DIODE \cite{vasiljevic2019diode}.
We report the results in Table \ref{tab:depth_comparison}. Our model achieves competitive performance compared to baselines that leverage pretrained text-to-image diffusion models, such as Marigold \cite{ke2024repurposing}. Notably, as illustrated in Figures \ref{fig:appendix_depth_comparison_1} and \ref{fig:appendix_depth_comparison_2}, our model demonstrates superior robustness than diffusion-based depth estimators like Marigold. Specifically, it excels in handling open-world images, including paintings, hazy weather, and unconventional textures.

\section{Conclusion}
\label{others}
Our experiments demonstrate that \algo achieves impressive results across a variety of tasks, including conditional T2I generation, depth estimation, open vocabulary semantic segmentation, pose estimation, multi-view generation, ID customization and camera pose estimation. We believe this work advances the capabilities of diffusion models, providing a versatile and scalable solution comparable to the flexibility offered by large language models. This represents a significant step toward developing a general-purpose vision model that can serve as the backbone for a wide variety of applications.

\section{Acknowledgements}
Stephan Mandt acknowledges support from the National Science Foundation (NSF) under an NSF CAREER Award IIS-2047418 and IIS-2007719, the NSF LEAP Center, by the Department of Energy under grant DE-SC0022331, the IARPA WRIVA program, the Hasso Plattner Research Center at UCI, the Chan Zuckerberg Initiative, and gifts from Qualcomm and Disney.

{
    \small
    \bibliographystyle{ieeenat_fullname}
    \bibliography{main}
}
\clearpage
\setcounter{page}{1}
\maketitlesupplementary

\section{Additional quantitative results}

\subsection{Camera Pose Estimation} 
We evaluate our model on camera pose estimation using the Google Scanned Object dataset \cite{downs2022google}. For this task, we use six rendered images of each synthetic object and estimate the camera poses by denoising the corresponding ray embeddings. Following RayDiffusion \citep{zhang2024cameras}, we apply least squares optimization to estimate the camera centers and rotations. The camera center accuracy, measured with a threshold of 0.3, is reported in Table \ref{table:poses}. 

\begin{figure}
    \centering
    \includegraphics[width=\linewidth]{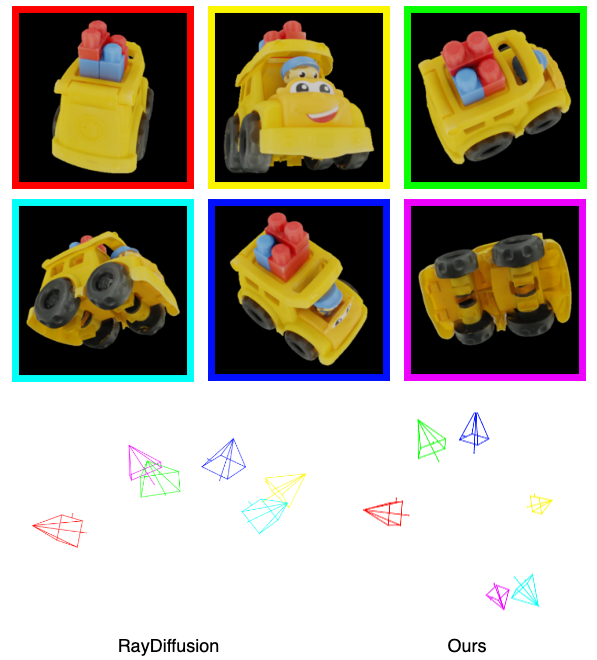}
    \caption{Qualitative comparison between RayDiffusion and \algo on GSO dataset. \algo yields better prediction.}
    \label{fig:cameraposes}
\end{figure}

Figure \ref{fig:cameraposes} provides a qualitative comparison between our model and RayDiffusion. RayDiffusion consistently predicts camera poses in the upper hemisphere due to the bias in its training data, such as CO3D, which predominantly features upper-hemisphere views. In contrast, thanks to the diversity of our large-scale training dataset, \algo achieves higher accuracy and avoids this limitation.

\begin{table}[t]
\centering
\resizebox{0.5\linewidth}{!}{%
\begin{tabular}{lccc}
\hline
Method & Accuracy \\
\hline
RayDiffusion \citep{zhang2024cameras} & 0.20 \\
\algo & 0.32 \\
\hline
\end{tabular}%
}
\caption{Comparison of zero-shot camera pose estimation methods on the GSO dataset, evaluated by Camera Center Accuracy at a threshold of 0.3.}
\vspace{-6pt}
\label{table:poses}
\end{table}

\subsection{Image Editing and Subject-driven generation}
We evaluate the performance of \algo for instruction-based image editing with the PIE-Bench dataset \cite{ju2023direct} in Table \ref{tab:pie_bench}; and for subject-driven generation using DreamBench \cite{ruiz2023dreambooth} in Table \ref{tab:dream_bench}. \algo achieves strong performance compared to specialized editing and generation approaches without any fine-tuning.

\begin{table}[t]
\centering
\footnotesize
\resizebox{\linewidth}{!}{%
\begin{tabular}{lcccccc}
\toprule
\textbf{Method} 
& \multicolumn{2}{c}{\textbf{Background Preservation}} 
& \multicolumn{2}{c}{\textbf{CLIP Semantics}} \\
\cmidrule(lr){2-3}\cmidrule(lr){4-5}
& \textbf{PSNR\,$\uparrow$} 
& \textbf{LPIPS\,$\downarrow$} 
& \textbf{Whole\,$\uparrow$} 
& \textbf{Edited\,$\uparrow$}\\
\midrule
Prompt-to-Prompt    & 17.87 & 208.80 & 25.01 & 22.44 \\
Null-text Inversion & 27.03 & 60.67  & 24.75 & 21.86 \\
PnPInversion        & 27.22 & 54.55  & 25.02 & 22.10 \\
Pix2pix-zero        & 20.44 & 172.22 & 22.80 & 20.54 \\
MasaCtrl            & 22.17 & 106.62 & 23.96 & 21.16 \\
InstructPix2Pix     & 16.69 & 271.33 & 23.49 & 22.20 \\
MGIE                & 21.20 & 142.25 & 24.28 & 21.79 \\
EditAR              & 21.32 & 117.15 & 24.87 & 21.87 \\
OneDiffusion        & 27.49 & 56.67  & 25.84 & 22.34 \\
\bottomrule
\end{tabular}
}
\caption{Evaluating image editing on PIE-Bench.}
\label{tab:pie_bench}
\end{table}

\begin{table}[t]
\centering
\resizebox{\linewidth}{!}{%
\begin{tabular}{lccc}
\toprule
\textbf{Methods} & \textbf{DINO}~$\uparrow$ & \textbf{CLIP-I}~$\uparrow$ & \textbf{CLIP-T}~$\uparrow$ \\
\midrule
Real Images (Oracle) & 0.774 & 0.885 & -- \\
\midrule
\multicolumn{4}{l}{\textit{Fine-Tuning}} \\
Textual Inversion & 0.569 & 0.780 & 0.255 \\
DreamBooth        & 0.668 & 0.803 & 0.305 \\
BLIP-Diffusion    & 0.670 & 0.805 & 0.302 \\
\midrule
\multicolumn{4}{l}{\textit{Tuning Free (Zero-shot)}} \\
Re-Imagen        & 0.600 & 0.740 & 0.270 \\
BLIP-Diffusion   & 0.594 & 0.779 & 0.300 \\
\algo & 0.692 & 0.814 & 0.297 \\
\bottomrule
\end{tabular}
}
\caption{Evaluating subject-driven generation on DreamBench.}
\label{tab:dream_bench}
\end{table}

\section{Additional qualitative results}

\begin{figure*}
    \centering
    \includegraphics[width=0.95\linewidth]{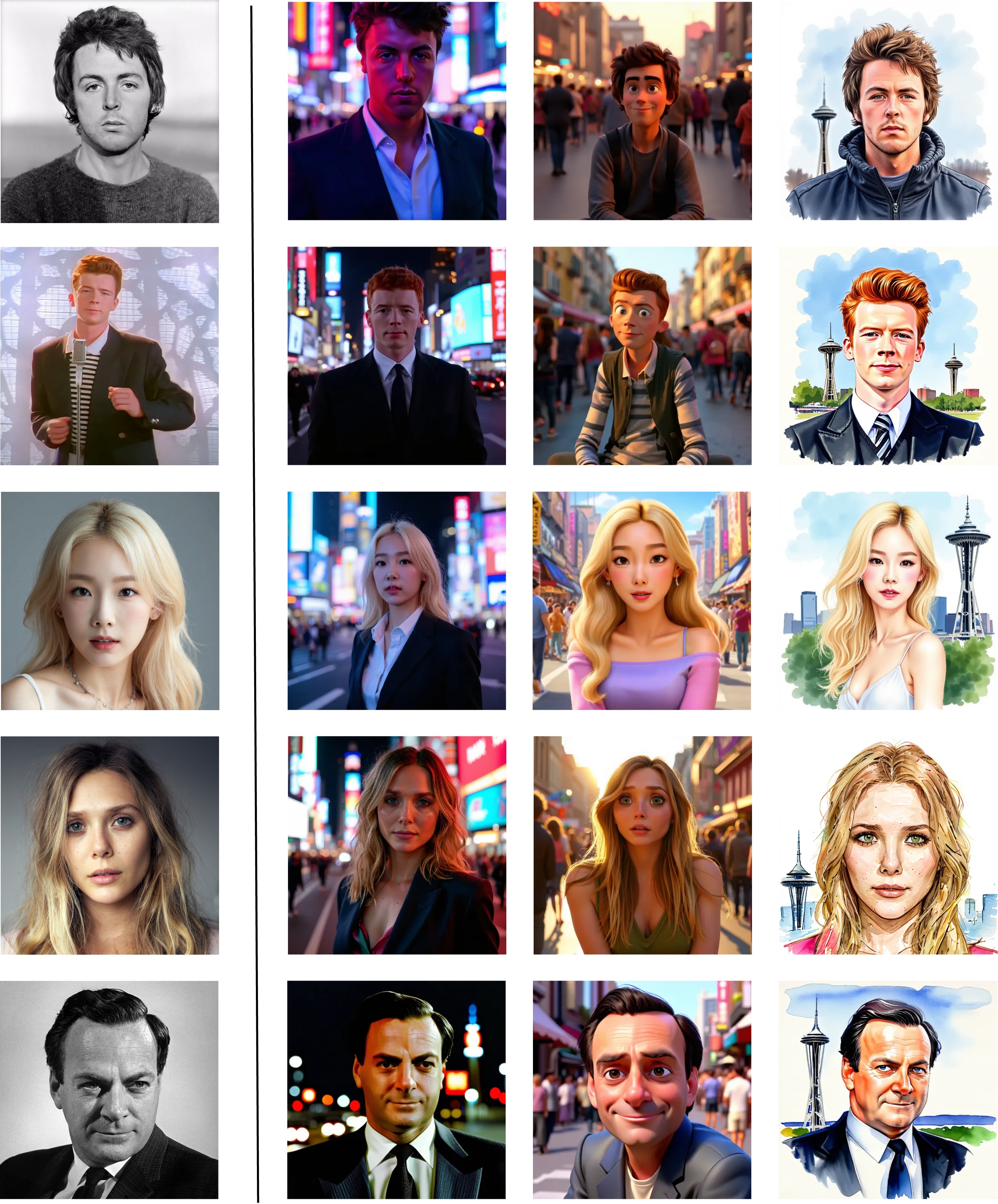}
    \caption{Qualitative results of \algo for (single reference) ID Customization task with photo of human faces. The left most images are input, target prompts for left to right columns are: 1) ``\emph{Photo of a man/woman wearing suit at Shibuya at night. He/She is looking at the camera}'', 2) ``\emph{pixarstyle, cartoon, a person in pixar style sitting on a crowded street}'', 3) ``\emph{watercolor drawing of a man/woman with Space Needle in background}''}
    \label{fig:appendix_faceid_demo_1}
\end{figure*}

\begin{figure*}
    \centering
    \includegraphics[width=0.9\linewidth]{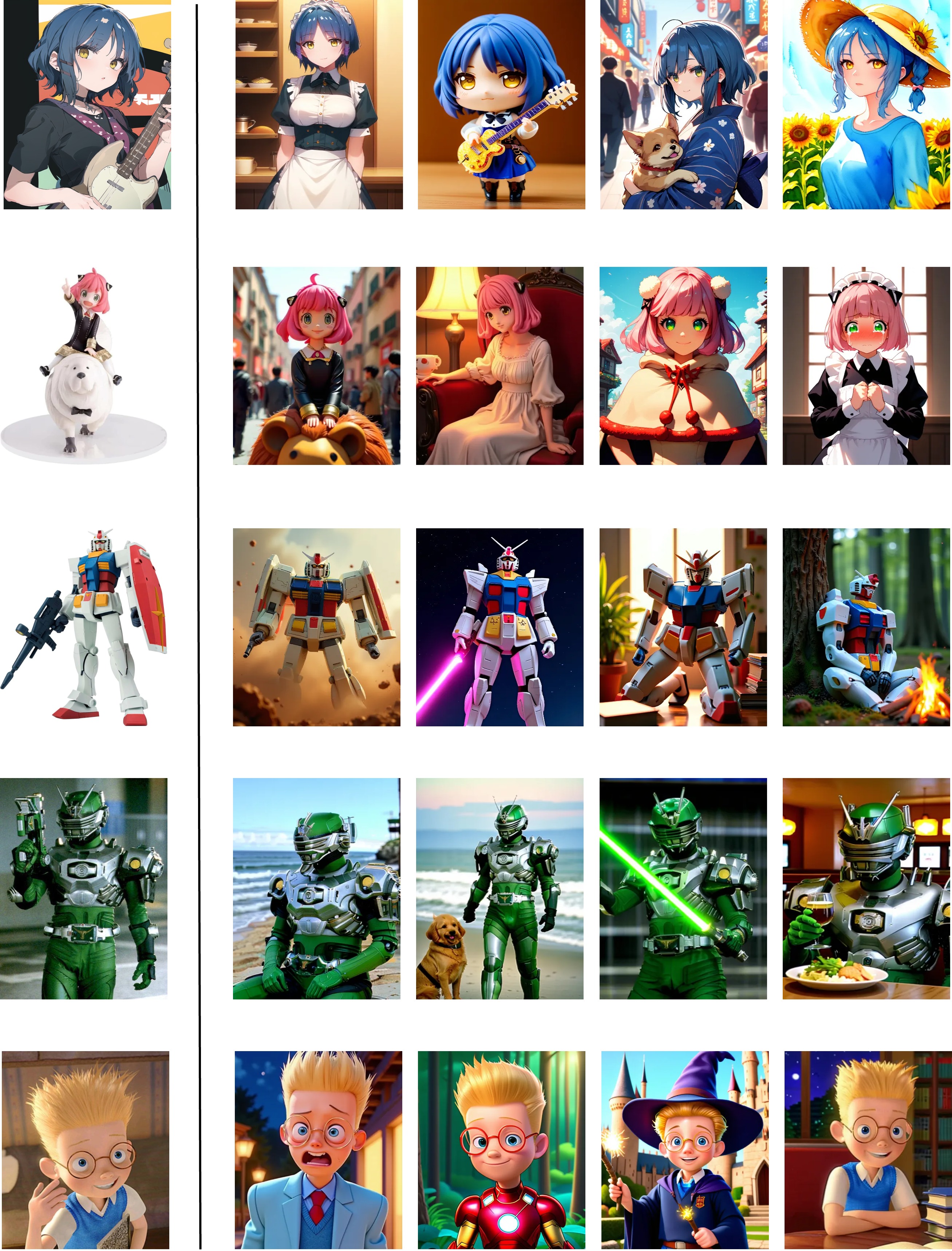}
    \vspace{-6pt}
    \caption{Qualitative results of \algo for (single reference) ID Customization task with photo of of non-human subjects or cartoon style input. \algo is highly versatile and can produce good results for all kind of input and not limited to photorealistic human images. Since we rely on attention, the model can attend to the condition view and preserve intricate details and is not limited by any bottleneck \eg latent representation.}
    \vspace{-8pt}
\label{fig:appendix_faceid_demo_2}
\end{figure*}

\paragraph{ID Customization.} We report ours results for ID Customization tasks in Figure \ref{fig:appendix_faceid_demo_1} and Figure \ref{fig:appendix_faceid_demo_2}. It can be observed from Figure \ref{fig:appendix_faceid_demo_1} that \algo well preserve the identity of a person with a single input and highly manipulatable. It can re-contextualize the image (as in first prompt), change the style from realistic photo to Pixar style (second prompt) and modify the medium to watercolor painting (third prompt). 

Moreover, our approach does not relying on face embedding as in previous works \cite{guo2024pulid, li2024photomaker, wang2024instantid} making it highly versatile. As illustrated in Figure \ref{fig:appendix_faceid_demo_2}, our model can preserver highly details, intricate structure as armor of the person in $4^{th}$ row. \algo can also work with non-human subject as the Gundam robot in $3^{rd}$ row. The model performs well with other style than photorealistic input such as anime ($1^{st}$ row), 3D figure ($2^{nd}$ row), cartoon ($5^{th}$ row). Our model is highly editable where we can control the style, human pose, camera angle, expression. 

\begin{figure*}
    \centering
    \includegraphics[width=0.9\linewidth]{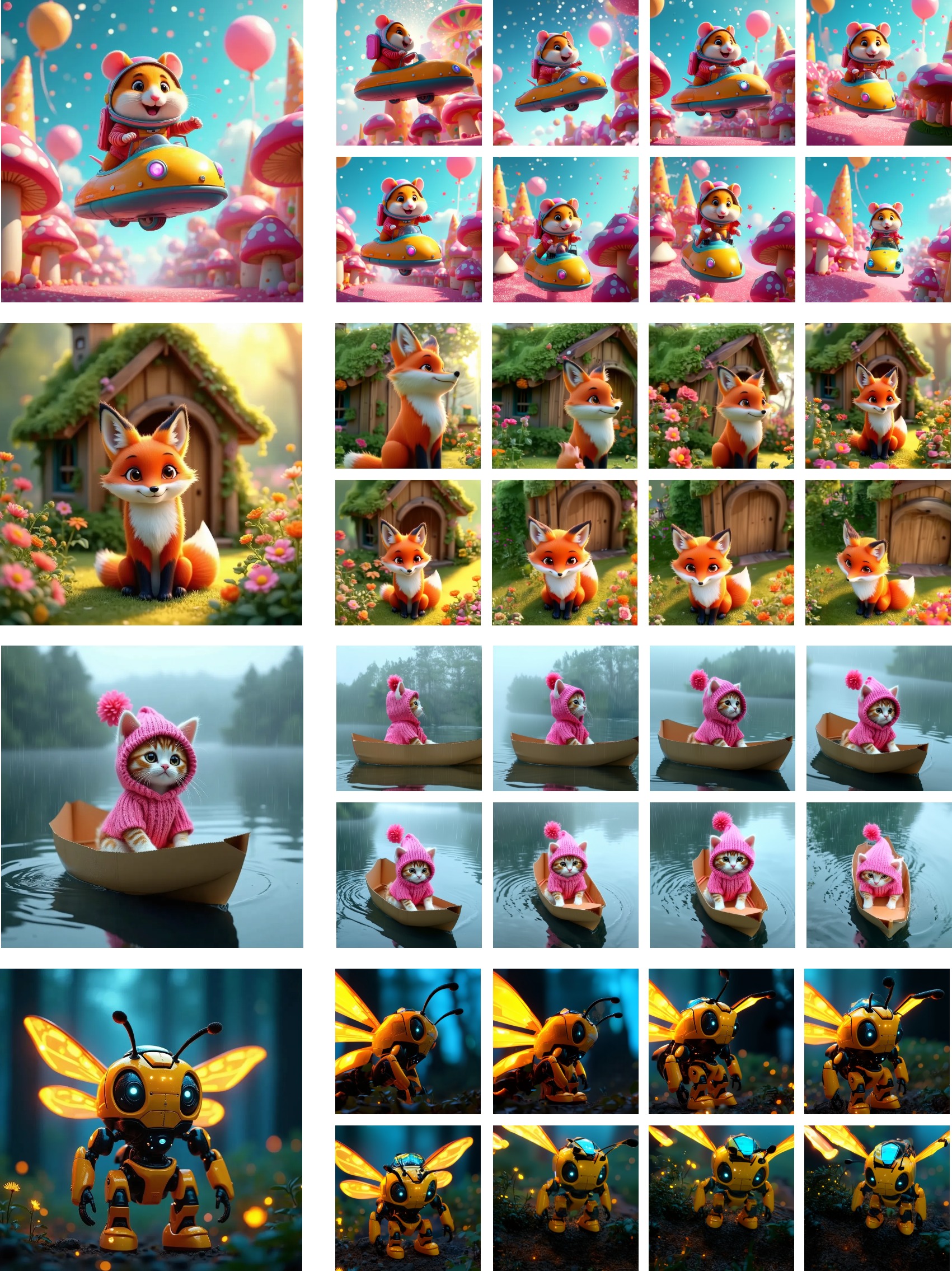}
    \caption{Qualitative results of image-to-multiview generation. The left most images are input. We equally slice the azimuth in range of $[-45, 60]$ and elevation in range of $[-15, 45]$ for all scenes.}
    \label{fig:appendix_multiview_demo_1}
\end{figure*}

\begin{figure*}
    \centering
    \includegraphics[width=0.9\linewidth]{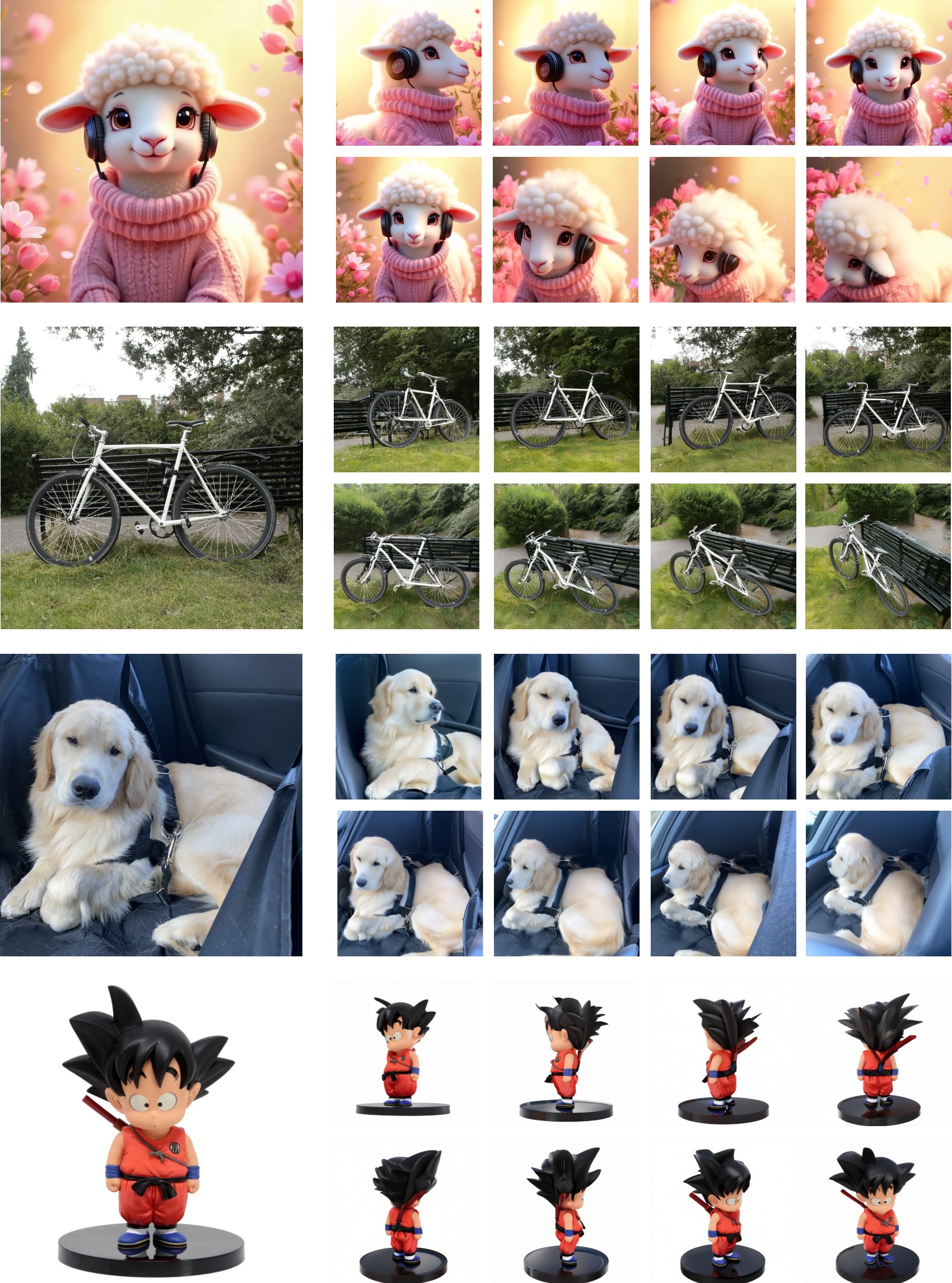}
    \caption{Qualitative results of image-to-multiview generation. We equally slice the azimuth in range of $[-45, 60]$ and elevation in range of $[-15, 45]$ for the first 3 scenes. For the last scene, the azimuth range is set to $[0; 360]$ and elevation range is set to $[-15; 15]$.}
    \label{fig:appendix_multiview_demo_2}
\end{figure*}

\vspace{-2mm}
\paragraph{Multiview generation} We report additional results for multiview generation in Figure \ref{fig:appendix_multiview_demo_1} and Figure \ref{fig:appendix_multiview_demo_2}. 
The generation process is as follow: we set the azimuth ranges to $[-0.45, 0.6]$ and elevation ranges to $[-15, 45]$, except for the last row of Figure \ref{fig:appendix_multiview_demo_2}. Then we equally slice these ranges to $80$ views. We first generate 3 anchor views from the input image and independently synthesize subsequent images based on input image and the nearest anchor. For each generation batch, we generate $3$ novels view and condition on $2$ images. We report views with index in $[0, 10, 20, \cdots, 70]$ in below figures. 

\algo is capable of generating photorealistic results of arbitrary objects or scenes from any number
of input views either realistic captured ($2^{nd}, 3^{rd}$ row of Figure \ref{fig:appendix_multiview_demo_2}) or synthesized images (Figure \ref{fig:appendix_multiview_demo_1}). Our model works best for camera trajectory covering front views of a scene. 

\begin{figure*}
    \centering
    \includegraphics[width=\linewidth]{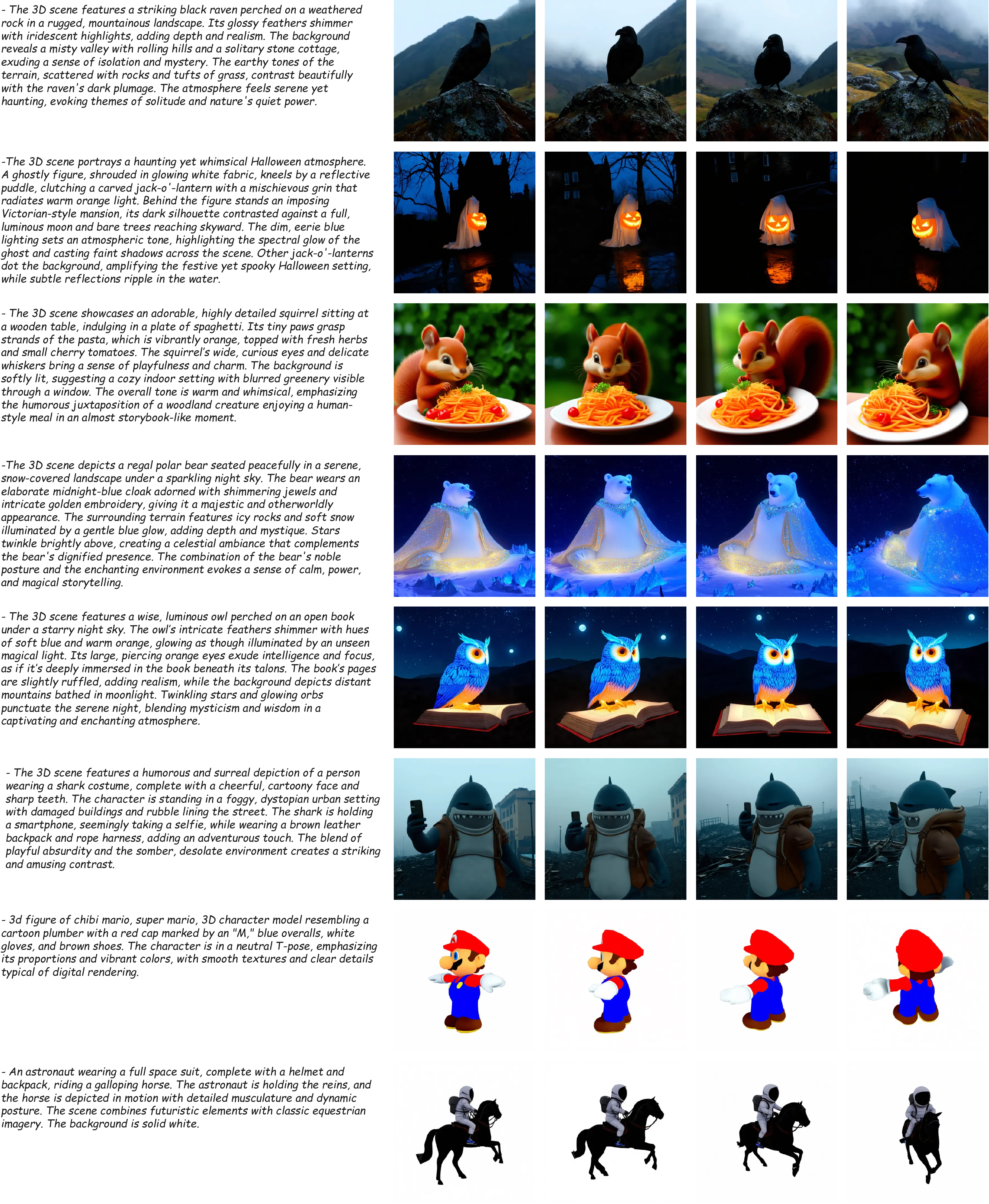}
    \caption{Qualitative results of text-to-multiview generation. The azimuth and elevation of left to right columns are $[0, 30, 60, 90]$ and $[0,10,20,30]$, respectively. We use following prefix for all prompts to improve the quality and realism of generated images: ``\emph{photorealistic, masterpiece, highly detail, score\_9, score\_8\_up}''. }
    \label{fig:appendix_text_to_multiview}
\end{figure*}

As mentioned ealier, \algo can also generate consistent multiview images from pure text and without any input images. Specifically, we simply input the azimuth and elevation as input for camera poses and generate all images from Gaussian noises as in Figure \ref{fig:appendix_text_to_multiview}. 

\begin{figure*}
    \centering
    \includegraphics[width=0.98\linewidth]{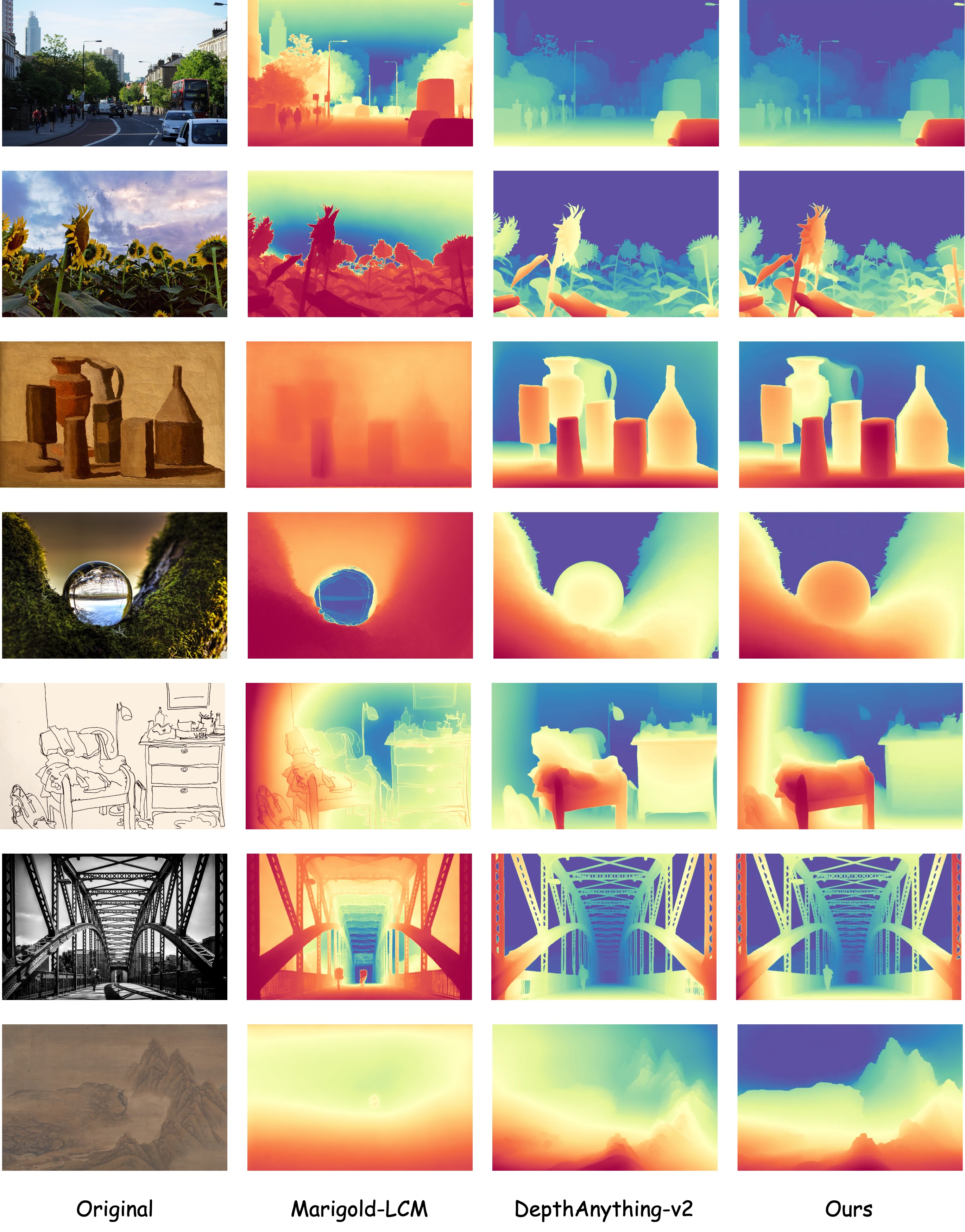}
    \vspace{-8pt}
    \caption{Qualitative comparison for depth estimation between \algo, Marigold \cite{ke2024repurposing} and DepthAnything-v2 \cite{yang2024depth}}
    \label{fig:appendix_depth_comparison_1}
\end{figure*}

\begin{figure*}
    \centering
    \includegraphics[width=0.98\linewidth]{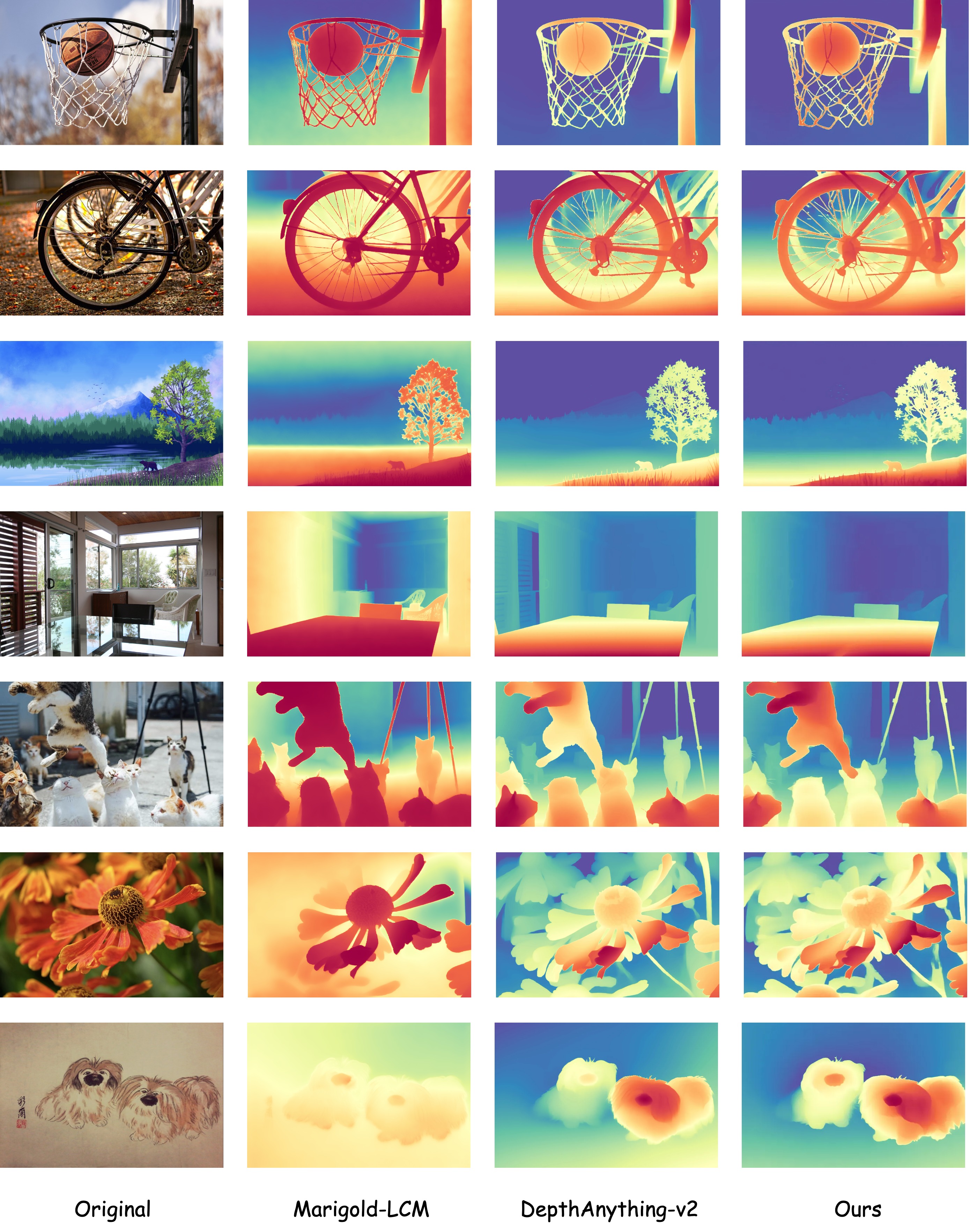}
    \vspace{-6pt}
    \caption{Qualitative comparison for depth estimation between \algo, Marigold \cite{ke2024repurposing} and DepthAnything-v2 \cite{yang2024depth}}
    \label{fig:appendix_depth_comparison_2}
\end{figure*}

\vspace{-2mm}
\paragraph{Depth estimation} We provide additional qualitative results of \algo and compare it with Marigold-LCM \cite{ke2024repurposing} and DepthAnything-v2 \cite{yang2024depth} in Figure \ref{fig:appendix_depth_comparison_1} and \ref{fig:appendix_depth_comparison_2}. We can see that our model estimator is more robust than Marigold on open-world test suits and is highly correlated with output of DepthAnything model. 

\begin{figure*}
    \centering
    \includegraphics[width=0.9\linewidth]{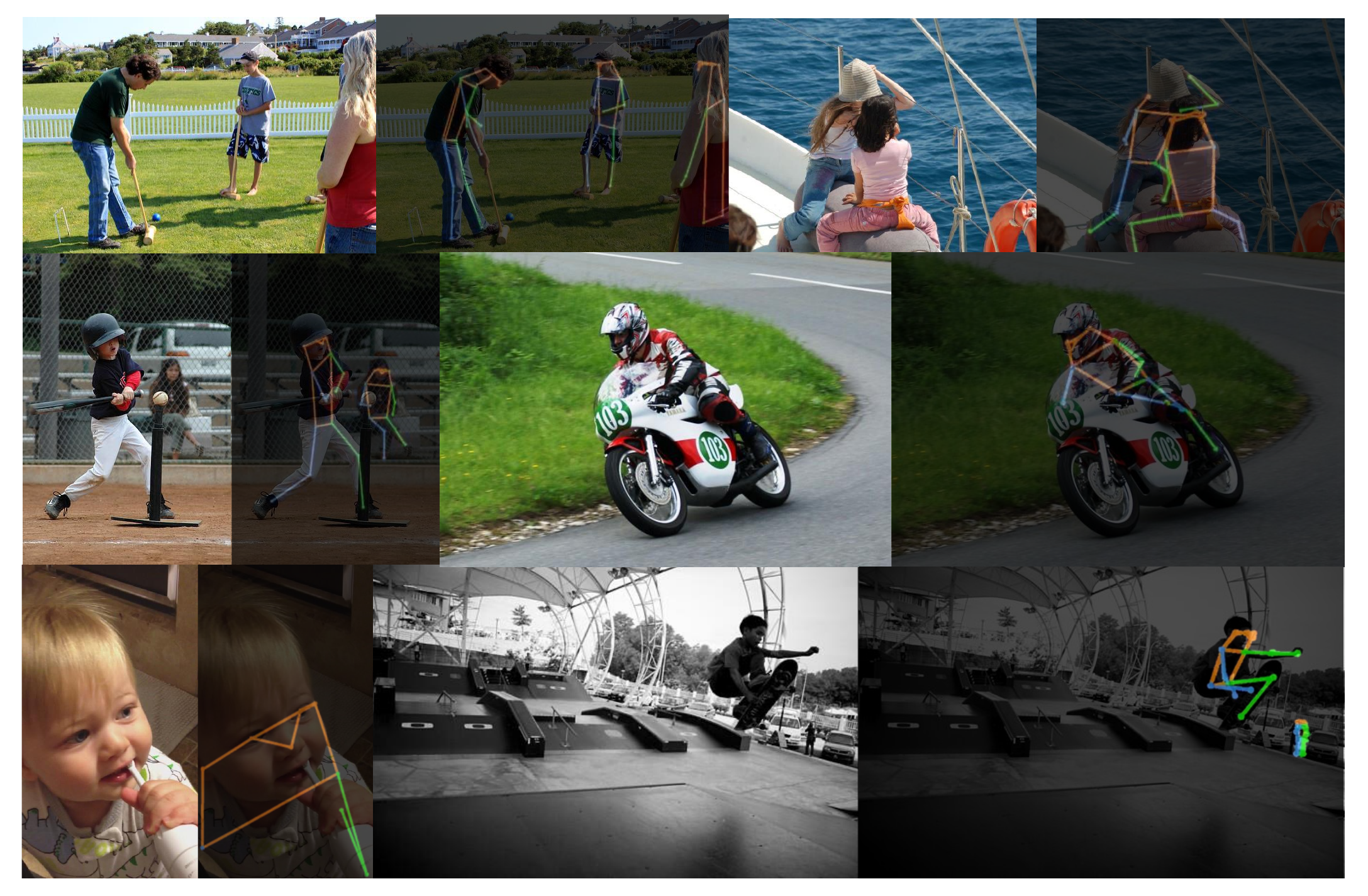}
    \vspace{-10pt}
    \caption{Qualitative examples of human pose estimation on COCO datasets.}
    \label{fig:appendix_pose_coco}
\end{figure*}

\vspace{-2mm}
\paragraph{Human Pose estimation} We report additional results for pose estimation on COCO dataset in Figure \ref{fig:appendix_pose_coco}. It can be observed that our model can predict multiple people in an image without relying on object detector models. 

\begin{figure*}
    \centering
    \includegraphics[width=0.9\linewidth]{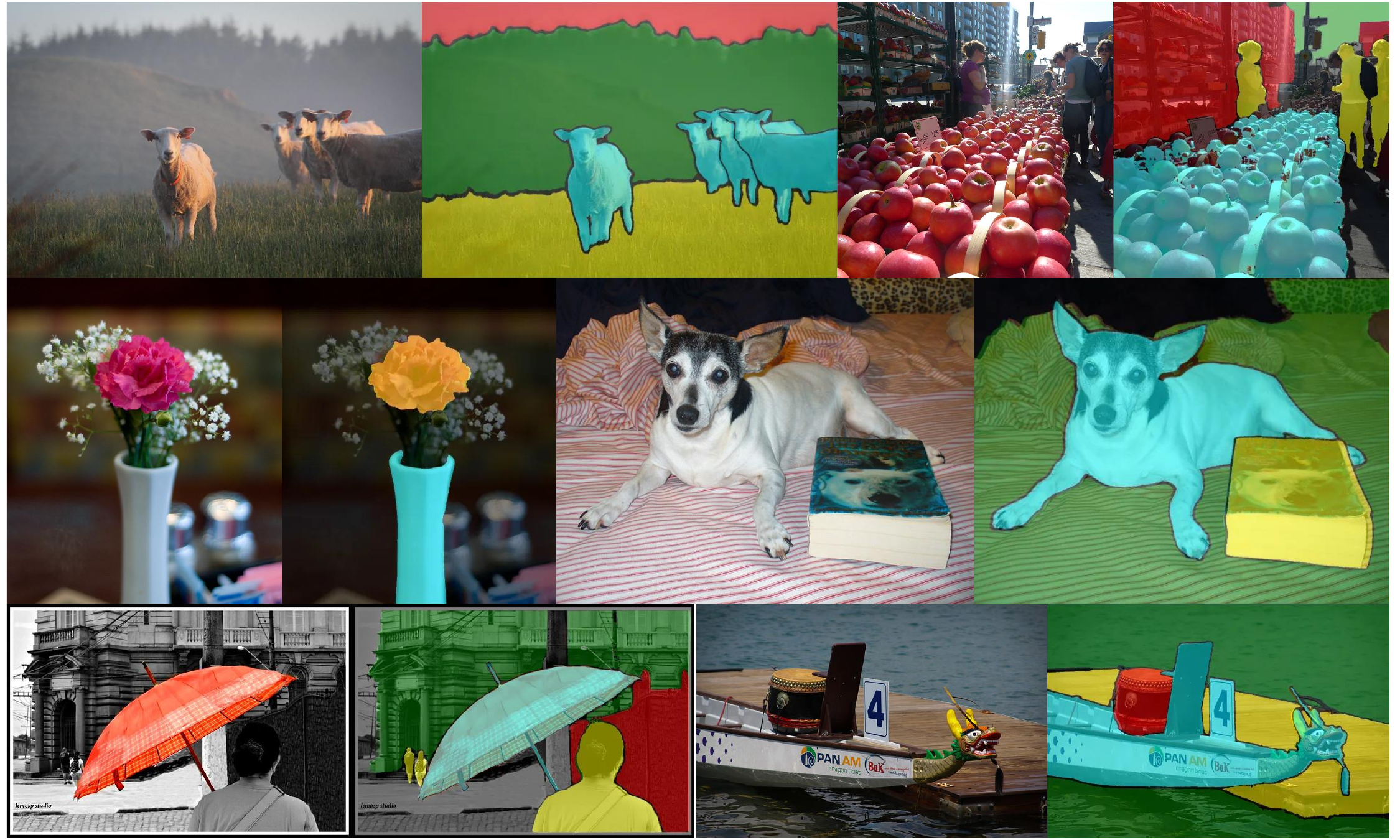}
    \caption{Qualitative examples of semantic segmentation on COCO datasets. The target class for each image (from left to right, from top to bottom) are 
    (\textcolor{cyan}{sheep}, \textcolor{darkyellow}{grass}, \textcolor{green}{mountain}, \textcolor{red}{sky}), 
    (\textcolor{cyan}{apple}, \textcolor{darkyellow}{person}, \textcolor{red}{building}), 
    (\textcolor{cyan}{vase}, \textcolor{darkyellow}{flower}, ), 
    (\textcolor{cyan}{dog}, \textcolor{darkyellow}{book}, \textcolor{green}{sheet}), 
    (\textcolor{cyan}{umbrella}, \textcolor{darkyellow}{person}, \textcolor{green}{building}, \textcolor{red}{gate}), 
    (\textcolor{cyan}{boat}, \textcolor{darkyellow}{dock}, \textcolor{red}{drum}).
    }
    \label{fig:appendix_seg_coco}
\end{figure*}

\vspace{-2mm}
\paragraph{Semantic Segmentation} We report qualitative results of semantic segmentation on COCO dataset in Figure \ref{fig:appendix_seg_coco}. Unlike previous models \cite{qin2023unicontrol, zhao2023uni}, our semantic-to-image and vice verse does not enforce hard association between colors and the target classes. We provide the color masks and class name as additional input in caption. 

\begin{figure}[ht]
  \centering
  \includegraphics[width=\linewidth]{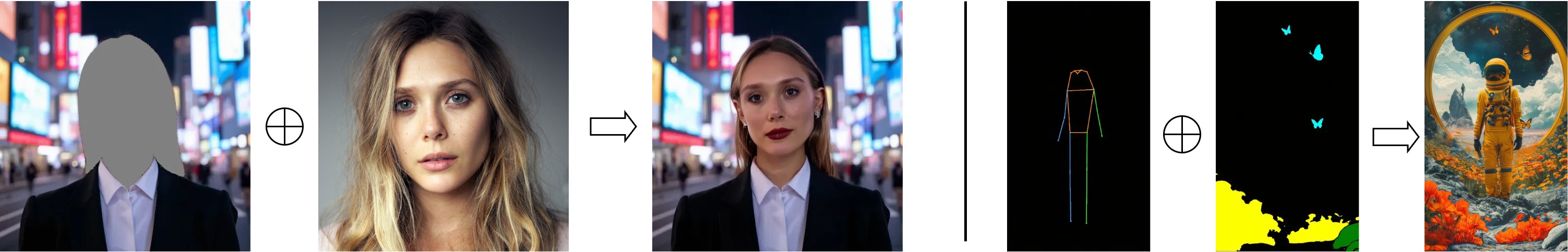}
   \caption{\algo is capable of performing several zero-shot task compositions. Left is inpainting with reference image and right is image generation with human pose and semantic map ({\color{cyan}butterfly}, {\color{darkyellow} flower}, {\color{misogreen} stone})}
   \label{fig:zeroshot}
\end{figure}

\vspace{-2mm}
\paragraph{Zero-shot task composition} \algo demonstrates remarkable generalization capabilities during inference, extending beyond its training on single-condition images to handle multiple conditioning inputs. Notably, it performs zero-shot task composition, such as inpainting with a reference face or generating images based on semantic maps and human pose, as illustrated in Figure \ref{fig:zeroshot}.

\section{Summary Datasets}
\begin{figure}[ht]
    \centering
    \includegraphics[width=0.8\columnwidth]{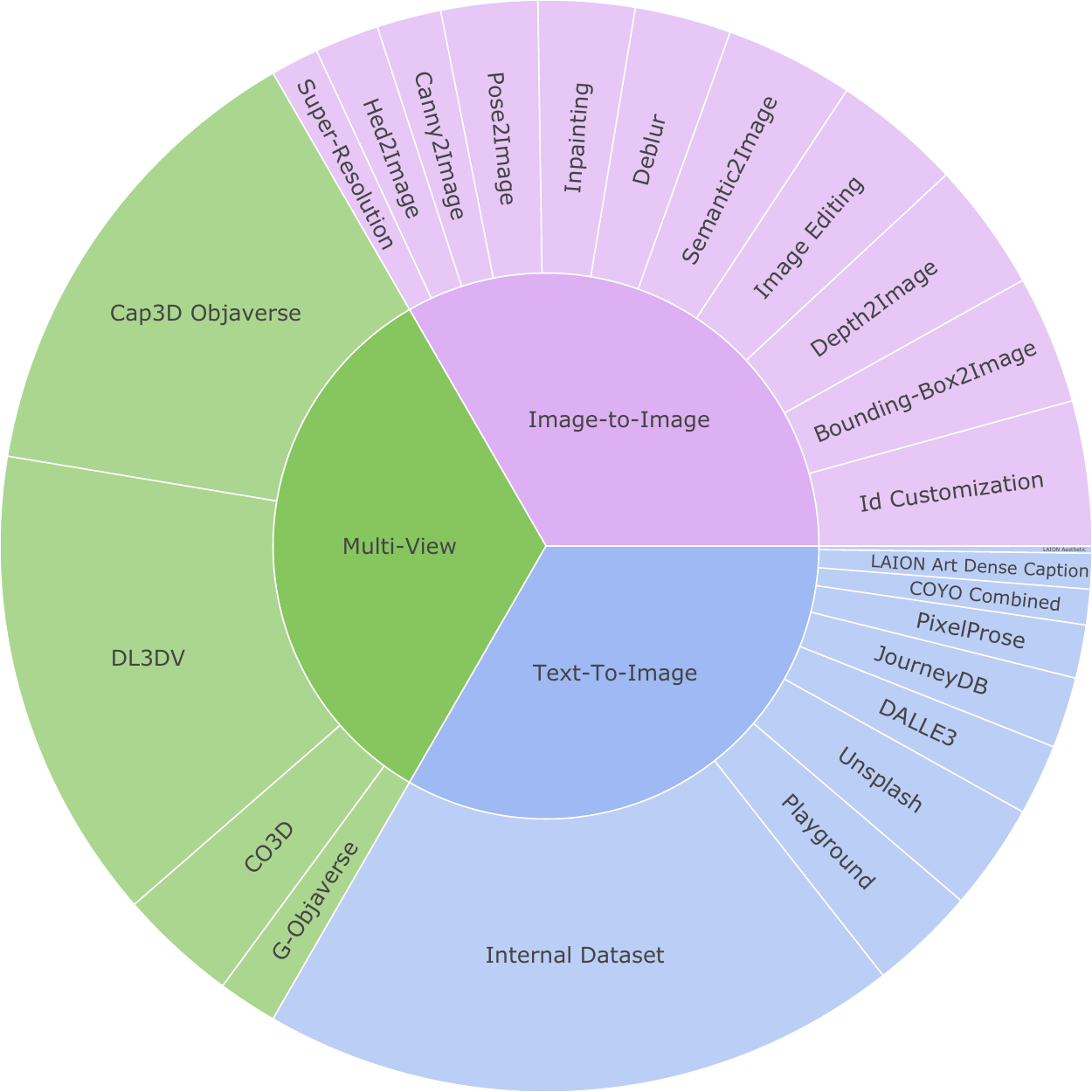}
    \caption{Distribution of training datasets for all tasks. Segments proportional to sampling rates. The inner section shows the super-category of target tasks, it can be observed that we train the model with equal budget for text-to-image, image-to-image and multiview generation. The outer section shows datasets used for each super-category.}
    \vspace{-10pt}
    \label{fig:mixture_rate_dataset}
\end{figure}
We train the model on multiple datasets reported in Section \ref{sec:experiment_dataset} and illustrated in Figure \ref{fig:mixture_rate_dataset}. The pie-chart segment each dataset proportional to the sampling rate of it in \textbf{third stage} of the training process. We train the model with equal budget for text-to-image, image-to-image translation ($2$ frames), and multiview generations ($2-6$ frames). Note that we filter and only use a subset of COYO with $11M$ images in our training. Due to the missing samples during download process, the LAION-aesthetic dataset only has $6M$ images. We recaption the LAION-aesthetic dataset with Molmo \cite{deitke2024molmo}.

\begin{figure*}
    \centering
    \includegraphics[width=0.9\linewidth]{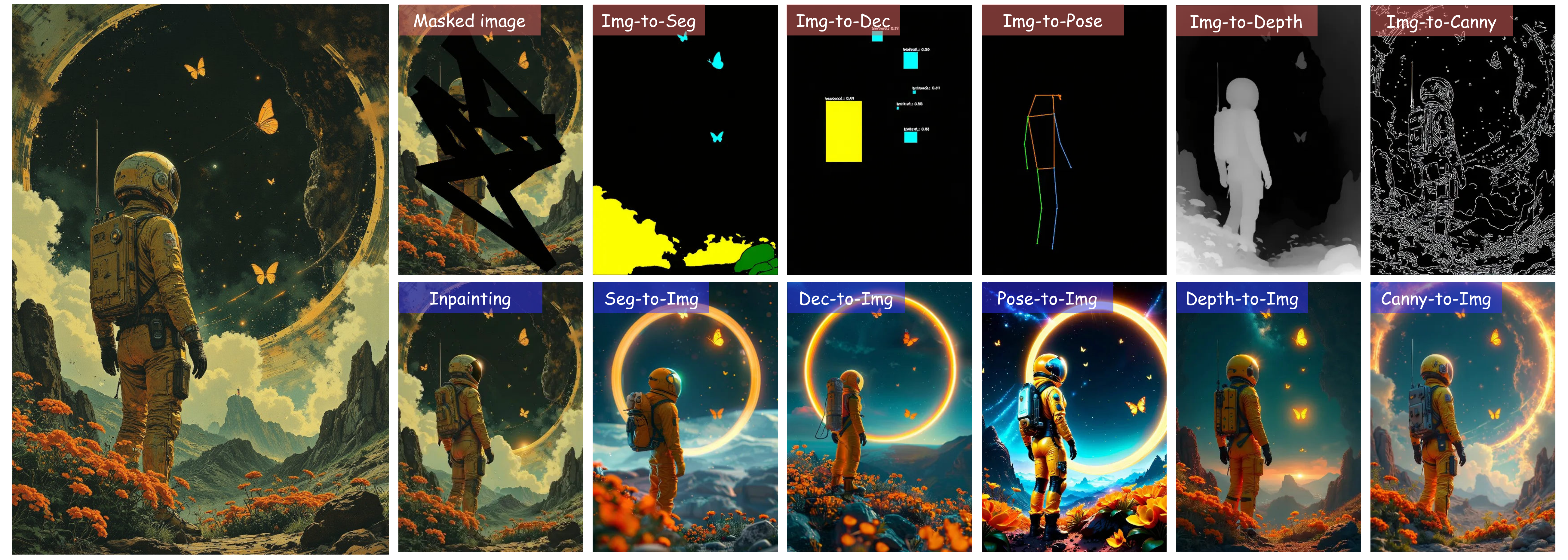}
    \caption{Illustration of our model capability to generate semantic mask, detection, human pose, depth, and canny edge from input image.
For semantic segmentation, we segment the flower (highlighted in yellow) and the rock (highlighted in green). For object detection, We localize the backpack (highlighted in yellow) and butterfly (highlighted in cyan).
Leveraging these conditions, we can reverse the process to recreate a variant of the input image based on the same caption.}
    \label{figure:appendix_bidirectional_examples}
\end{figure*}

\end{document}